\documentclass{article}

\usepackage[preprint]{neurips_2023}

\usepackage[utf8]{inputenc} 
\usepackage[T1]{fontenc}    
\usepackage{url}            
\usepackage{booktabs}       
\usepackage{amsfonts}       
\usepackage{amsmath}
\usepackage{amssymb}
\usepackage{nicefrac}       
\usepackage{microtype}      
\usepackage[dvipsnames]{xcolor}         
\usepackage[colorlinks=true, allcolors=Blue]{hyperref}
 \usepackage{natbib}
\usepackage{algorithm}
\usepackage{algpseudocode}
\usepackage{graphicx}
\usepackage{subcaption}
\usepackage{multirow}
\usepackage{float}
\usepackage[mathlines]{lineno}

\title{Estimating Reaction Barriers\\ with Deep Reinforcement Learning}

\author{%
  Adittya Pal \\
  Institut for Matematik og Datalogi, Syddansk Universitet \\
  Campusvej 55, 5230 Odense M, Danmark \\
  \texttt{adpal@imada.sdu.dk} \\
}

\begin{document}

\maketitle

\begin{abstract}
Stable states in complex systems correspond to local minima on the associated potential energy surface. Transitions between these local minima govern the dynamics of such systems. Precisely determining the transition pathways in complex and high-dimensional systems is challenging because these transitions are rare events, and isolating the relevant species in experiments is difficult. Most of the time, the system remains near a local minimum, with rare, large fluctuations leading to transitions between minima. The probability of such transitions decreases exponentially with the height of the energy barrier, making the system's dynamics highly sensitive to the calculated energy barriers. This work aims to formulate the problem of finding the minimum energy barrier between two stable states in the system's state space as a cost-minimization problem. It is proposed to solve this problem using reinforcement learning algorithms. The exploratory nature of reinforcement learning agents enables efficient sampling and determination of the minimum energy barrier for transitions.
\end{abstract}

\section{Introduction}

There are multiple sequential decision-making processes that one comes across in the world, such as control of robots, autonomous driving, and so on. Instead of constructing an algorithm from the bottom up for an agent to solve these tasks, it would be much easier if one could specify the environment and the state in which the task is considered solved and let the agent learn a policy that solves the task \cite{rl_survey, rl_agent}. Reinforcement learning attempts to address this problem. It is a hands-off approach that provides a feature vector representing the environment and a reward for the actions the agent takes \cite{sutton}. The objective of the agent is to learn the sequence of steps that maximizes the sum of returns. \cite{reward_rl}

\begin{figure*}[h!]
    \centering
    \begin{subfigure}[t]{0.49\textwidth}
        \includegraphics[width=0.8\textwidth, angle= -90]{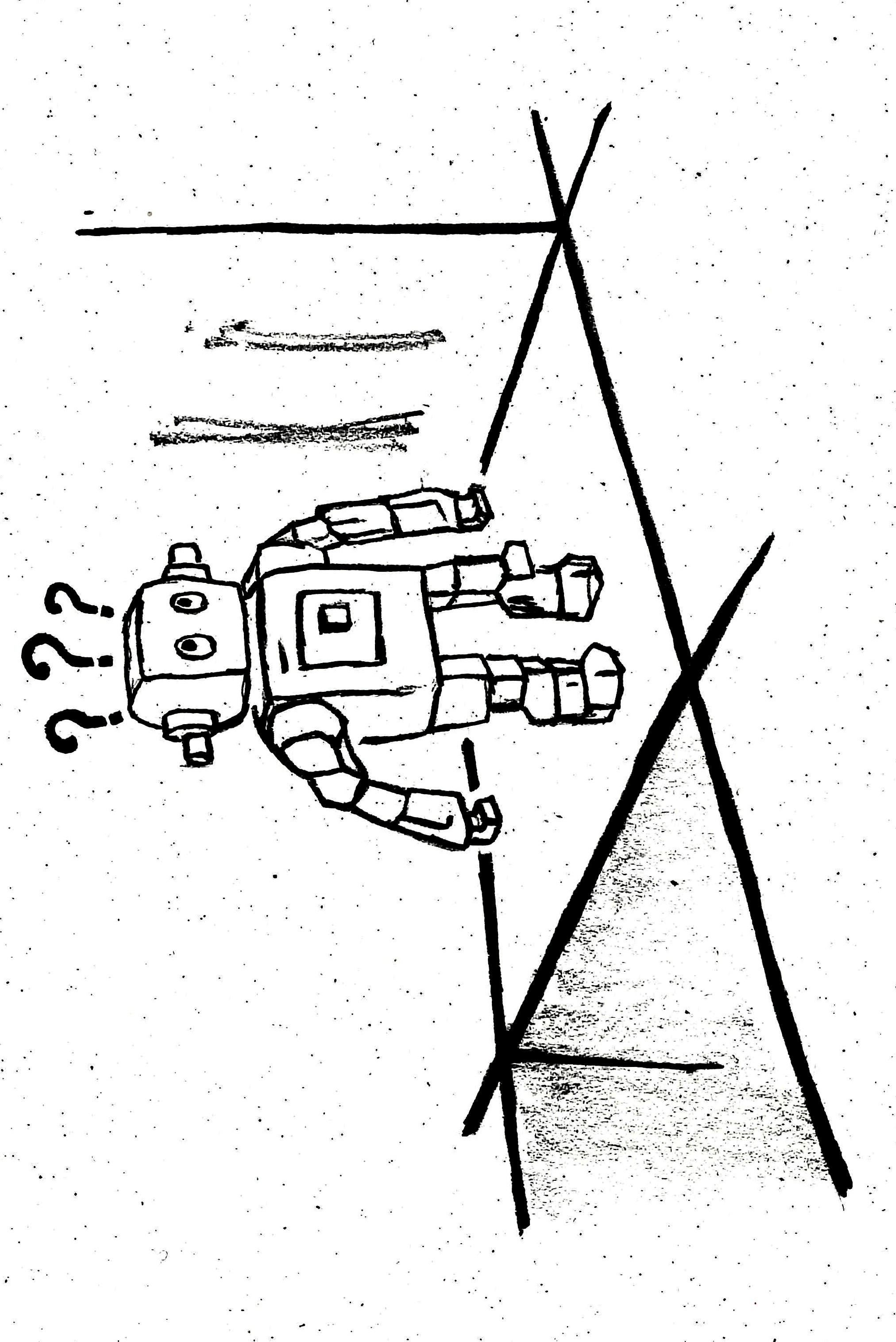}
        \caption{}
    \end{subfigure}%
    ~ 
    \begin{subfigure}[t]{0.49\textwidth}
        \centering
        \includegraphics[width=0.8\textwidth, angle= -90]{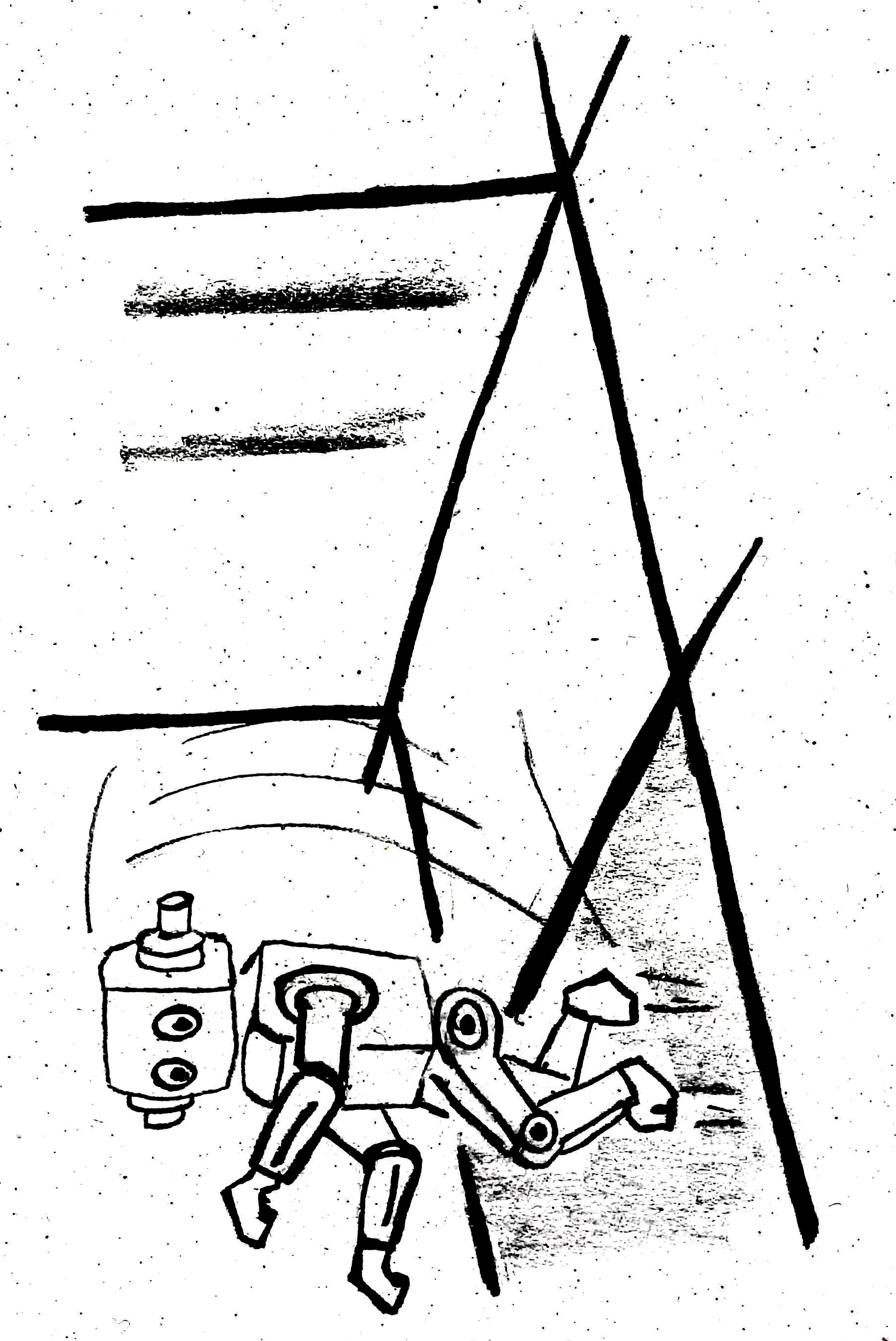}
        \caption{}
    \end{subfigure}
    \begin{subfigure}[t]{0.49\textwidth}
        \includegraphics[width=0.8\textwidth, angle= -90]{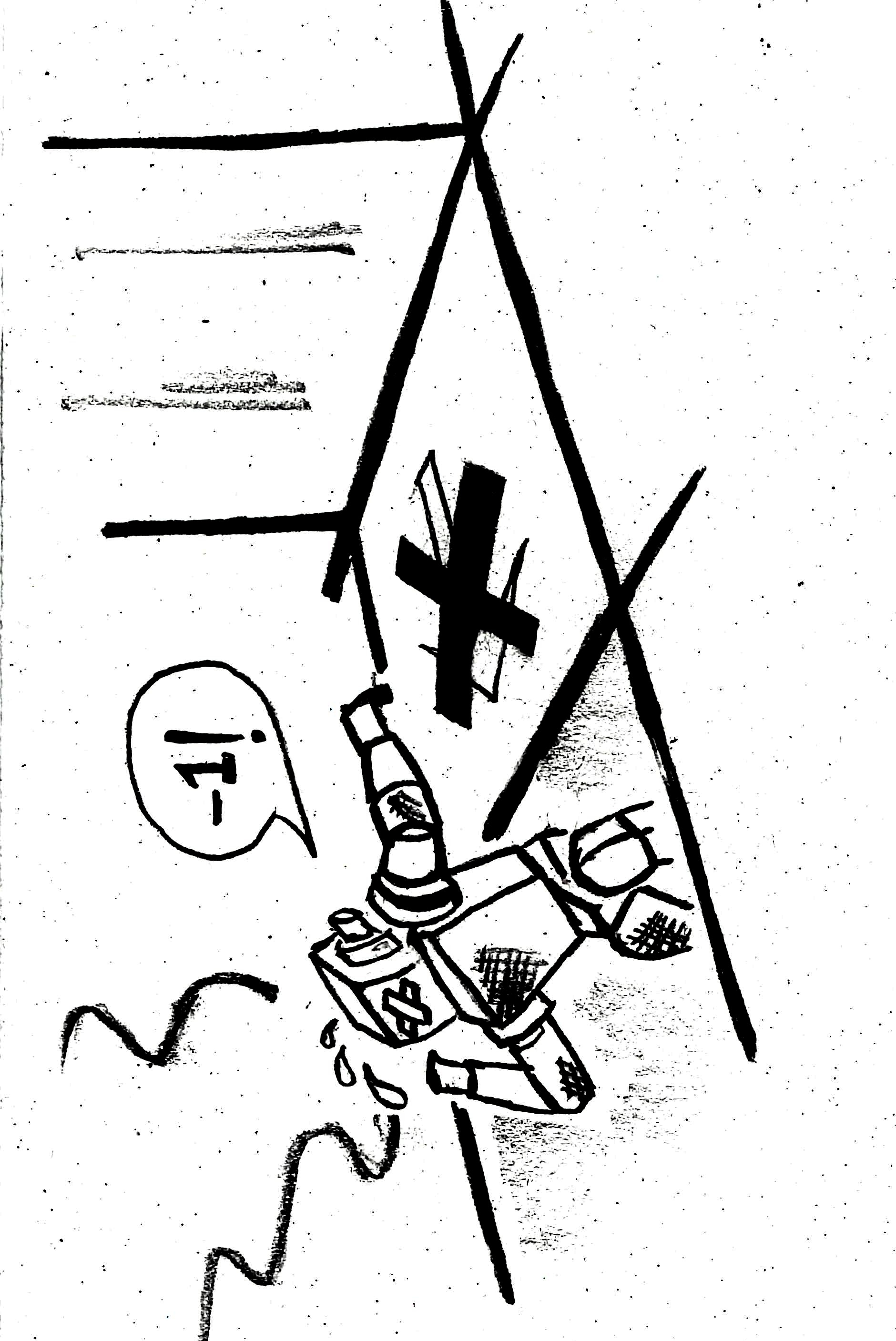}
        \caption{}
    \end{subfigure}%
    ~ 
    \begin{subfigure}[t]{0.49\textwidth}
        \centering
        \includegraphics[width=0.8\textwidth, angle= -90]{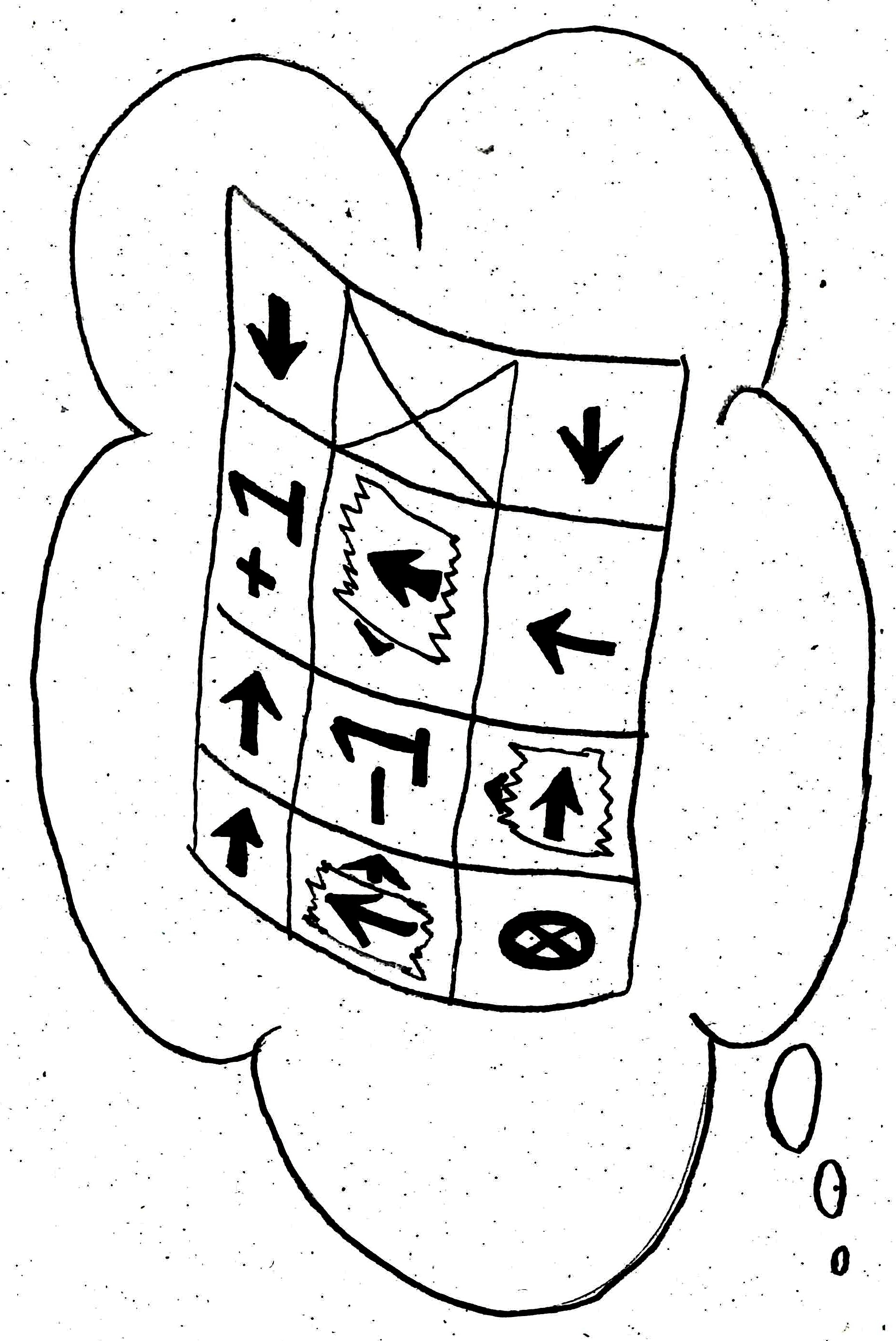}
        \caption{}
    \end{subfigure}
    \caption{Maze solving using reinforcement learning: (a) The agent is at a state at a particular time step, and takes an action to reaches the next state (b). The agent records the reward obtained by taking the action in that state and (c) continues exploring the environment. After a large number of interactions with the environment, the agent learns a policy (d) that maximizes the rewards collected by the agent. The policy (d) gives the sequence of actions that the agent has to take from the initial state to the final state so that it collects the maximum rewards in an episode.}
    \label{maze_rl}
\end{figure*}

One widespread example of a sequential decision-making process in which reinforcement learning is utilized is solving mazes \cite{maze_rl, maze_rl2}. The agent, a maze runner, selects a sequence of actions that might have long-term consequences \cite{maze_rl3}. Since the consequences of immediate actions might be delayed, the agent must evaluate the actions it chooses and learn to select actions that solve the maze. Particularly, in the case of mazes, it might be relevant to sacrifice immediate rewards for possibly larger rewards in the long term. This is the exploitation-exploration trade-off, where the agent has to learn to choose between leveraging its current knowledge to maximize its current gains or further increasing its knowledge for a potential larger reward in the long term, possibly at the expense of short-term rewards \cite{map_maze_rl, neural_map}. The process of learning by an agent while solving a maze is illustrated in Figure \ref{maze_rl}.

GridWorld is an environment for reinforcement learning that mimics a maze \cite{sutton}. The agent is placed at the start position in a maze with blocked cells, and the agent tries to reach a stop position with the minimum number of steps possible. One might note an analogy of a maze runner with an agent negotiating the potential energy landscape of a transition event for a system along the saddle point with the minimum height. The start state and the stop state are energy minima on the potential energy surface, separated by an energy barrier for the transition. The agent would have to perform a series of perturbations to the system to take it from one minimum (the start state) to another (the end state) through the located saddle point. In the case of a maze, the (often discrete) action changes the position of the agent (by a fixed measure), but while locating minimum energy pathways, the physical problem demands a continuous action space. However, as in the case of a maze, an action changes the variables describing the system (be it physical coordinates or state variables) by a small measure. 

As in the maze-solving problem, the agent tries to identify the pathway with the minimum energy barrier. If the number of steps is considered to be the cost incurred in a normal maze, then it is the energy along the pathway that is the cost of the transition event. Reward maps can vary depending on the maze considered, but in the original GridWorld problem, the agent was given a negative reward if the action led to a wall cell, and a zero reward for all non-terminal states (a discount factor $<1$ enforces the minimum count of steps). In the case of locating trajectories with a low energy barrier, the agent should be penalized if the action leads to a state with progressively higher energies (but not to an extent that it hinders exploration). The exact reward map used is detailed later in Section \ref{methods}.

\begin{figure*}
    \centering
    \begin{subfigure}[t]{0.49\textwidth}
        \includegraphics[width=\textwidth]{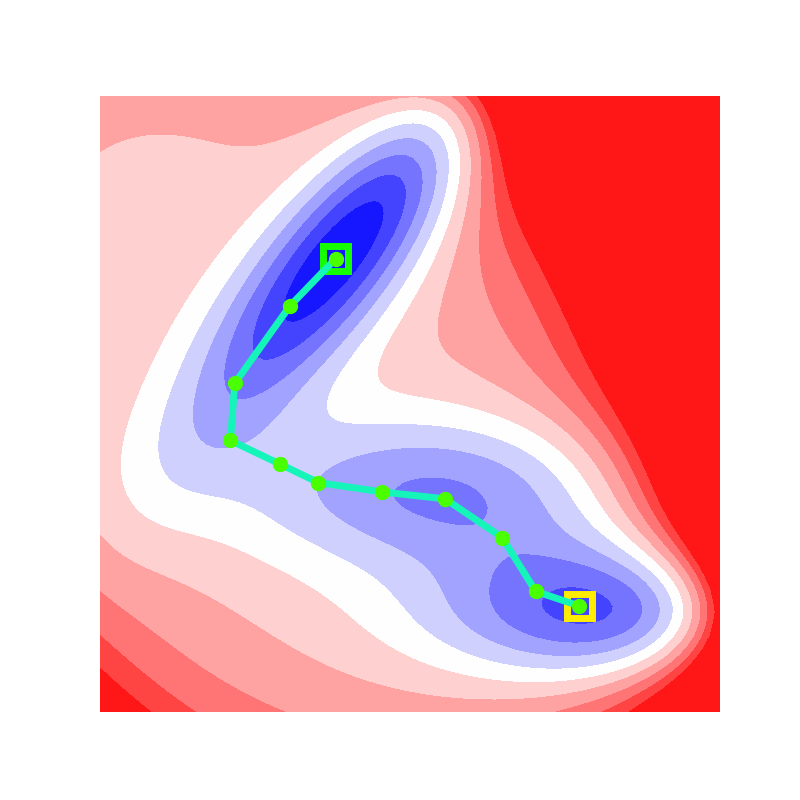}
        \caption{}
    \end{subfigure}%
    \hfill
    \begin{subfigure}[t]{0.45\textwidth}
        \centering
        \includegraphics[width=\textwidth]{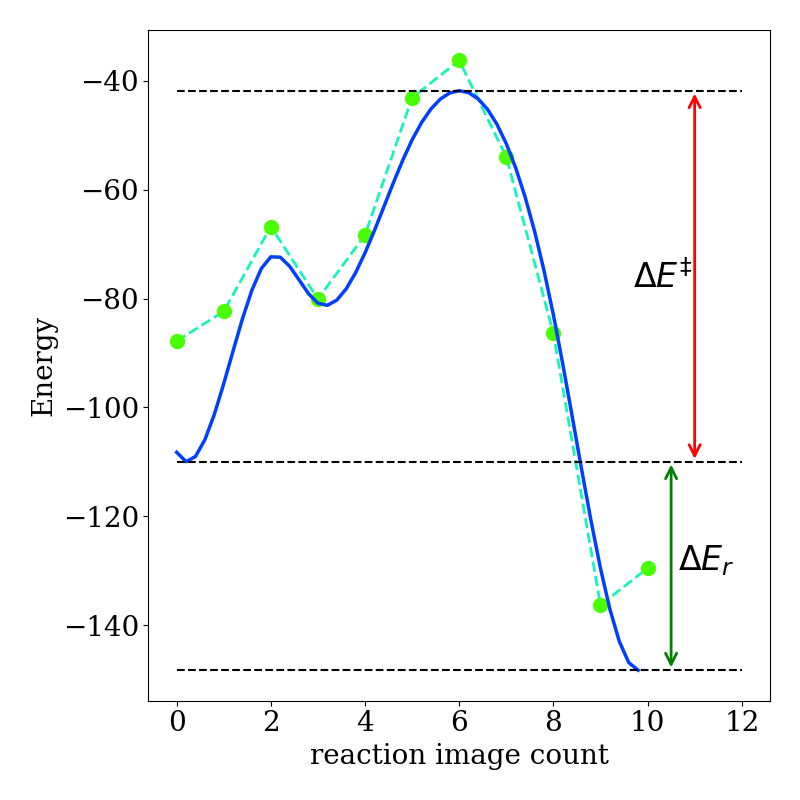}
        \caption{}
    \end{subfigure}
    \begin{subfigure}[t]{0.49\textwidth}
        \includegraphics[width=\textwidth]{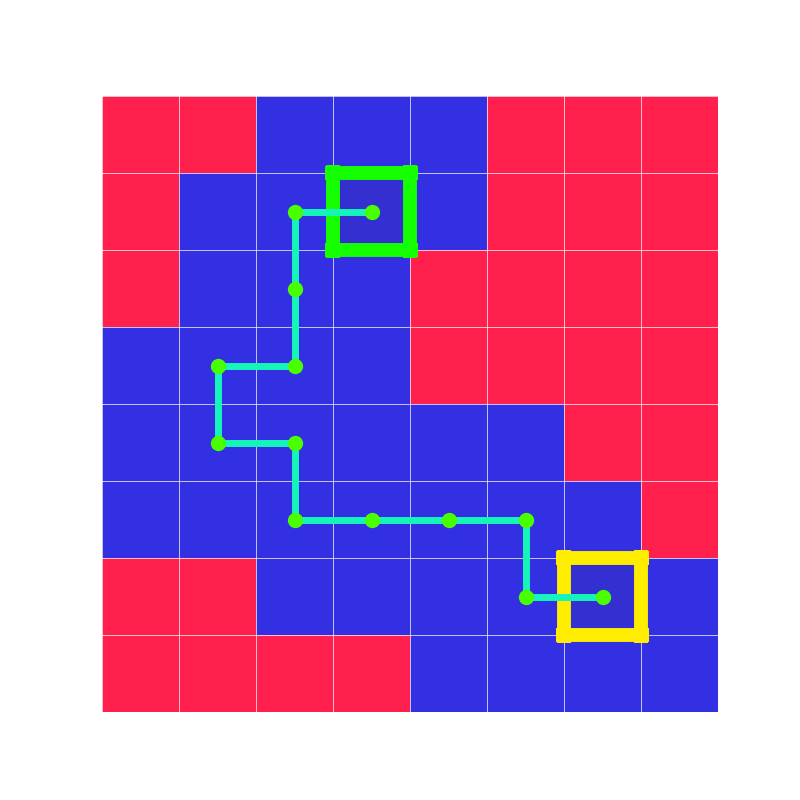}
    \end{subfigure}%
    ~ 
    \begin{subfigure}[t]{0.49\textwidth}
        \centering
        \includegraphics[width=\textwidth]{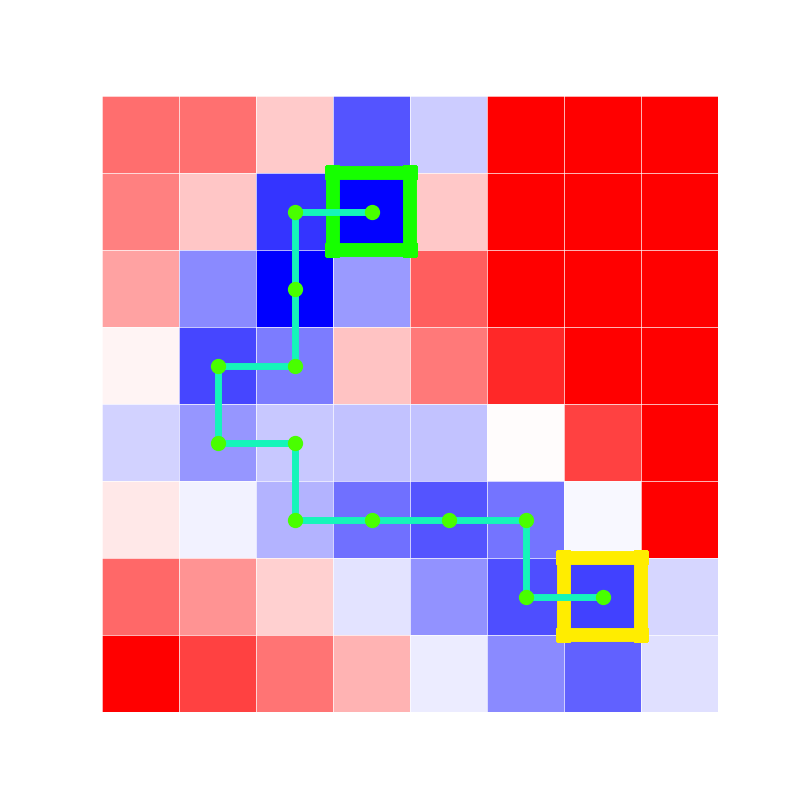}
        \caption{}
    \end{subfigure}
    \caption{Estimating reaction barriers by modeling the potential energy surface as a maze: (a) The pathway with the lowest energy barrier as determined by a growing string method on the potential energy surface with 9 intermediate images. (b) The reaction profile, plotted as a solid blue line (interpolated to give a smooth curve) from the pathway determined by the growing string method. The reaction barrier is marked as $\Delta E^\ddagger$. Instead of the extreme binary classification of a grid cell as a wall or move as in the maze (c), each cell can be assigned an energy value as in (d).}
    \label{pes_rl}
\end{figure*}

A comparison is attempted in Figure \ref{pes_rl}. A smooth potential energy surface is coarse-grained to construct a maze, where all positions with a negative potential energy are shaded blue (possible move cells), while those with a positive potential energy are shaded red (representing walls). The initial state in the maze is marked yellow, while the final state is marked green. However, instead of classifying a grid cell of the constructed maze either as a wall or as a cell, one can discretize the state space and assign an energy value to a cell. An agent can then be trained to reach the final state starting from the initial state and collect the maximum sum of rewards along its path (minimizing the energy along the pathways requires assigning the negative of the energy as the reward for an action leading the agent to the cell). Since this is an episodic problem, one already runs into the problem where the agent moves back and forth between two adjacent cells, collecting rewards from each move in an attempt to maximize the sum of rewards collected, rather than reaching the final state and terminating the episode. For this simple setting, the problem is solved by rewarding the agent only the first time it visits a cell and terminating the episode after a fixed number of steps (in this case, $15$). The energy profile of the pathway followed by the agent (inferred from the rewards collected in an episode) in this maze is plotted as the dashed green line in Figure \ref{pes_rl}b. As can be seen, coarse-graining the potential energy surface into an $8\times 8$ maze and then solving it using standard reinforcement learning algorithms provides a reasonable starting point to address the problem.

The problem of locating the minimum energy barrier for a transition has applications in physical phase transitions, synthesis plans for materials, activation energies for chemical reactions, and the conformational changes in biomolecules that lead to reactions inside cells. In most of these scenarios, the dynamics are governed by the kinetics of the system (rather than the thermodynamics) because the thermal energy of the system is much smaller than the energy barrier of the transition. This leads to the system spending most of its time around the minima, and some random large fluctuations in the system lead to a transition. This is precisely why transition events are rare and difficult to isolate and characterize with experimental methods. Moreover, these ultra-fast techniques can be applied to only a limited number of systems. Because transition events are rare, sampling them using Monte Carlo methods requires long simulation times, making them inefficient \cite{mc_inefficient}. To sample the regions of the potential energy surface around the saddle point adequately, a large number of samples have to be drawn. Previous work has been done to identify the saddle point and determine the height of the transition barrier---transition path sampling \cite{tps}, nudged elastic band \cite{neb}, growing string method \cite{mgsm}, to name a few---which use ideas from gradient descent. However, even for comparatively simple reactions, these methods are not always guaranteed to find the path with the energy barrier that is a global minimum because the initial guess for the pathway might be wrong and lead to a local minimum.

With the advent of deep learning and the use of neural nets as function approximators for complex mappings, there has been increased interest in the use of machine learning \cite{rare_event_sampling} to either guess the configuration of the saddle point along the pathway (whose energy can then be determined by standard \emph{ab initio} methods) or directly determine the height of the energy barrier given the two endpoints of the transition. Graph neural networks \cite{gnn}, generative adversarial networks \cite{gan}, gated recurrent neural networks \cite{grnn}, transformers \cite{transformer}, machine-learned potentials \cite{ml_potential1, ml_potential2}, and so on, have been used to optimize the pathway for such transitions.

Noting the superficial similarities between solving a maze and determining the transition pathway with the lowest energy barrier, it is proposed to use standard and tested deep reinforcement learning algorithms used to solve mazes in an attempt to solve the problem of finding minimum energy pathways. The problem is formulated as a min-cost optimization problem in the state space of the system. An actor function approximator suggests the action to be taken by the agent when it is in a particular state. A critic function approximator provides an estimate of the sum of rewards until the end of the episode from the new state after taking the action suggested by the actor. Actor-critic based reinforcement learning techniques have been shown to solve problems effectively, even in higher dimensions \cite{ac_high_dim}.  Neural nets are used as the actor and critic function approximators, and a randomly perturbed policy is used to facilitate exploration of the potential energy surface by the agent. Delayed policy updates and target policy averaging are used to stabilize learning, especially during the first few epochs, which are crucial to the optimal performance of the agent. This formulation is used to determine the barrier height of the optimal pathway in the M{\"u}ller-Brown potential.

Section \ref{methods} describes the methods used to formulate the problem as a Markov decision process and the algorithm used to solve it. Section \ref{results} elaborates on the experiments in which the formulated method is used to determine the barrier height of a transition on the M{\"u}ller-Brown potential. Section \ref{discussion} contains a short discussion of the work in the context of other similar studies, while Section \ref{conclusion} outlines the conclusions drawn from this work.

\section{Methods}
\label{methods}

To solve the problem of finding a pathway with the lowest energy barrier for a transition using reinforcement learning, one has to model it as a Markov decision process. Any Markov decision process consists of (state, action, next state) tuples. In this case, the agent starts at the initial state (a local minimum) and perturbs the system (action) to reach a new state. Since the initial state was an energy minimum, the current state will have a higher energy. However, as in many sequential control problems, the reward is delayed. A series of perturbations that lead to states with higher energies might enable the agent to climb out of the local minimum into another one that contains the final state. By defining a suitable reward function and allowing the agent to explore the potential energy surface, it is expected that the agent will learn a path from the initial state to the final state that maximizes the rewards. If the reward function is defined properly, it should correspond to the pathway with the lowest energy barrier for the transition.

Once the problem is formulated as a Markov decision process, it can be solved by some reinforcement learning algorithm. In most reinforcement learning algorithms,  this (state, action, reward, next state, next action) tuple is stored while the agent is learning. Twin Delayed Deep Deterministic Policy Gradient (TD3) \cite{td3} is a good start because it prevents overestimation of the state value function, which often leads the agent to exploit the errors in the value function and learn a suboptimal policy. Soft Actor Critic (SAC) \cite{sac1} tries to blend the deterministic policy gradient with a stochastic policy optimization, promoting exploration by the agent. In practice, using a stochastic policy to tune exploration often accelerates the agent's learning.

\subsection{Markov Decision Process}
\label{mdp}

The Markov decision process is defined on:
\begin{itemize}
\item a state space $\mathcal{S}$, consisting of states $s \in \mathbb{R}^d$, where $d$ is the dimensionality of the system, chosen to be the number of degrees of freedom in the system.
\item a continuous action space $\mathcal{A}$, where each action $\Delta s \in \mathbb{R}^d: |\left(\Delta s\right)_i| \leq 1$ is normalized, and the action is scaled using an appropriate scaling factor $\lambda$.
\end{itemize}
In a state $s^{(k)}$, the agent takes an action $\Delta s^{(k)}$. Since the action is considered a perturbation of the current state of the system, the next state $s^{(k+1)}$ is determined from the current state $s^{(k)}$ as $s^{(k+1)} = s^{(k)} + \lambda\cdot \Delta s^{(k)}$.

To determine the minimum energy barrier for a transition, the reward for an action taking the agent to state $s^{(k+1)}$ from state $s^{(k)}$ is chosen to be the negative of the energy of the next state, $-E\left(s^{(k+1)}\right)$. The negation makes maximizing the sum of rewards collected by the reinforcement learning agent in an episode equivalent to minimizing the sum of energies along the pathway for the transition. The reward acts as immediate feedback to the agent for taking an action in a particular state. However, what is important is the long-term reward, captured by the sum of the rewards over the entire episode, leading the agent to identify a transition pathway with a low sum of energies at all intermediate steps.

Since both the state space and action space are continuous, an actor-critic based method, specifically the soft actor-critic (SAC), is used. Additionally, since the state space is continuous, the episode is deemed to have terminated when the difference between the current state and the target state is smaller than some tolerance, $x \in \mathbb{R}^d : |x - x_t| < \delta$ for some small $\delta$. Otherwise, it would be extremely unlikely that the agent would land exactly at the coordinates of the final state after taking some action. An obvious problem with this definition of the Markov process is that the agent may prefer to remain in a state near the target state (but far enough so that the episode does not terminate), collecting rewards for the rest of the episode. This behavior was observed in Section \ref{results}.

\subsection{Algorithm}
\begin{algorithm}
    \caption{Computing minimum energy barrier using SAC in environment \texttt{env}}
    \begin{algorithmic}[1]
        \State Initialize actor net parameters $\theta$ and critic Q-net parameters $\phi_1$, $\phi_2$
        \State Hard update target Q-net parameters: $\phi_{i, \text{target}} \leftarrow \phi_i$ for $i = 1, 2$
        \State Initialize replay buffer $\mathcal{R}$
        \For{\texttt{step = 1} to \texttt{numEpisodes}}
            \State \texttt{state, \_ = env.reset()}
            \For{\texttt{t = 0} to \texttt{maxSteps}}
                \State let actor select an action by policy $\pi_\theta$
                \State \textcolor{Blue}{perturb the action with some noise $\epsilon \sim \mathcal{N}(0, \sigma)$ : $a^{(k)} = \pi_\theta(s^{(k)}) + \text{clip}(\epsilon, -\epsilon_{\text{lim}}, \epsilon_{\text{lim}})$}
                \State execute \texttt{action} in the \texttt{env} and observe the $\left(s^{(k)}, a^{(k)}, r^{(k)}, s^{(t+1)}, \text{\texttt{ done}}\right)$ tuple
                \State push it to the replay buffer $\mathcal{R}$
                \If {\texttt{step \% agent\_update == 0 }}
                \State sample a minibatch $\mathcal{B}$ of $\left(s, a, r, s^\prime, \text{\texttt{ done}}\right)$ tuples from the replay buffer $\mathcal{R}$
                \State compute targets as \textcolor{Blue}{(where $a^\prime = \pi_\theta(\cdot|s^\prime) + \text{clip}(\epsilon, -\epsilon_{\text{lim}}, \epsilon_{\text{lim}})$)}
                \begin{equation*}
                     \text{\hspace{2cm}}y(r, s^\prime) = r + \gamma (1 - \text{\texttt{done}}) \left(\min_{i=1,2} Q_{\phi_{i, \text{target}}}(s^\prime, a^\prime) - \alpha\log \pi_\theta(a^\prime|s^\prime)) \right)
                \end{equation*}
                \State \parbox[t]{\dimexpr\textwidth-\leftmargin-\labelsep-\labelwidth}{update critic Q-nets parameters by one step of gradient descent with loss function \\\textcolor{Blue}{(with the gradient clipped by some maximum value)}}
                \begin{equation*}
                    \text{\hspace{2cm}}\frac{1}{|\mathcal{B}|}\,\,\,\nabla_{\phi_i}\left(\sum_{s\in\mathcal{B}} \min_{i=1,2} Q_{\phi_i}(s, a) -  y(r, s^\prime)\right)^2 \text{for i = 1, 2}
                \end{equation*}
                \If {\textcolor{Blue}{\texttt{t \% update\_target == 0}}}
                    \State \parbox[t]{\dimexpr\textwidth-\leftmargin-\labelsep-\labelwidth}{update actor net parameters by one step of gradient descent with loss function \\\textcolor{Blue}{(with the gradient clipped by some maximum value)}}
                    \begin{equation*}
                        \text{\hspace{2.5cm}}\frac{1}{|\mathcal{B}|}\,\,\,\nabla_\theta\left(\sum_{s\in\mathcal{B}} \min_{i=1,2} Q_{\phi_i}(s, \pi_\theta(s)) - \alpha\log \pi_\theta(a|s) \right)^2
                    \end{equation*}
                    \State update the entropy coefficient $\alpha$ as one step of gradient descent with loss function
                    \begin{equation*}
                        \text{\hspace{2.5cm}}\frac{1}{|\mathcal{B}|}\,\,\,\nabla_\alpha\left(\sum_{s\in\mathcal{B}} - \alpha\log \pi_\theta(a|s) - \alpha\log a^\prime \right)^2
                    \end{equation*}
                    \State soft update the target networks: $\phi_{i, \text{target}} \leftarrow \tau\phi_i + (1-\tau)\phi_{i, \text{target}}$
                \EndIf
                \EndIf
            \EndFor
        \EndFor
        \State return the actor net parameters $\theta$ and critic Q-net parameters $\phi_1, \phi_2$.
\end{algorithmic}
\label{used_algorithm}
\end{algorithm}

SAC, an off-policy learning algorithm with entropy regularization, is used to solve the formulated Markov Decision process because the inherent stochasticity in its policy facilitates exploration by the agent. Entropy regularization tries to balance maximizing the returns till the end of the episode with randomness in the policy driving the agent. The algorithm learns a behavior policy $\pi_\theta$ and two critic Q-functions, which are neural networks with parameters $\phi_1$ and $\phi_2$ (line 1 of Algorithm \ref{used_algorithm}).

The agent chooses an action $a^{(k)} \equiv \Delta s^{(k)}$ to take when at state $s^{(k)}$ following the policy $\pi_\theta$ (line 8). 
Returns from state $s^{(k)}$ when acting according to policy $\pi$ is the discounted sum of rewards collected from that step onward to the end of the episode: $R_t = - \sum_{i = t}^T \gamma^{i-t}\,E(s^{(i)})$. 
The objective of the reinforcement learning agent is to determine the policy $\pi^*$ that maximizes the returns $R_{t}$, for states $s \in \mathcal{S}$.
This is done by defining a state-action value function, $Q(s^{(i)}, a^{(i)})$, which gives an estimate of the expected returns if the agent takes action $a^{(i)}$ when in state $s^{(i)}$: $Q(s^{(i)}, a^{(i)}) = \mathbb{E}\left[R_t : s_t = s^{(i)}, a_t = a^{(i)}\right]$. Since the objective is to maximize the sum of the returns, the action-value function can be recursively defined as

\begin{equation*}
    Q(s^{(i)}, a^{(i)}) = - E(s^{(i+1)}) + \gamma\,\max_{a^{(i+1)} \in \mathcal{A}} Q(s^{(i+1)}, a^{(i+1)})
\end{equation*}
which is implemented in line 14 of Algorithm \ref{used_algorithm}.

A replay buffer with a sufficiently large capacity is employed to increase the probability that independent and identically distributed samples are used to update the actor and two critic networks. The replay buffer (in line 3) is modeled as a deque where the first samples to be enqueued (which are the oldest) are also dequeued first, once the replay buffer has reached its capacity and new samples have to be added. Since an off-policy algorithm is used, the critic net parameters are updated by sampling a mini-batch from the replay buffer at each update step (line 13). Stochastic gradient descent is used to train the actor and the two critic nets.

The entropy coefficient $\alpha$ is adjusted over the course of training to encourage the agent to explore more when required and to exploit its knowledge at other times (line 18) \cite{sac2}. However, some elements from the TD3 algorithm \cite{td3} are borrowed to improve the learning of the agent, namely delayed policy updates and target policy smoothing. Due to the delayed policy updates, the critic Q-nets are updated more frequently than the actor and the target Q-nets to allow the critic to learn faster and provide more precise estimates of the returns from the current state. To address the problem of instability in learning, especially in the first few episodes while training the agent, target critic nets are used. Initially, the critic nets are duplicated (line 2), and subsequently soft updates of these target nets are carried out after an interval of a certain number of steps (line 19). This provides more precise estimates for the state-action value function while computing the returns for a particular state in line 14.
Adding noise to input during the training of a machine learning model often leads to improved performance because it acts as an $L_2$-regularizer \cite{noise_regularize} and prevents the model from memorizing patterns in the training data sample in supervised learning scenarios. Similarly, TD3 adds noise to the action predicted by the actor network to smooth out the Q-function, so that the agent does not memorize the imprecise estimates of the Q-function early on during the training process. This prevents the policy from exploiting the imprecise estimates of the Q-function approximator for certain actions, reducing the chances of learning a brittle policy that does not generalize well. The logic is that, for well-behaved, smooth reward maps, the reward should not abruptly change with small differences in the action.
The addition of clipped noise to the action chosen by the actor net (in line 9) also encourages the agent to explore the potential energy surface. The changes to the SAC algorithm, borrowed from TD3, are highlighted in blue in the pseudocode of Algorithm \ref{used_algorithm}. The parameters used in the particular implementation of the algorithm are listed in Table \ref{tab: parameters}.

\begin{table}    
    \centering    
    \begin{tabular}{c | r c}\hline
        & Parameter & Value \\\hline
        \multirow{4}{*}{$Q_\phi(s,a)$} & network architecture & 4-256-256-1 \\
        & activation for hidden layer & $relu$ \\
        & activation for output layer & none \\
        & learning rate & $10^{-4}$ \\\hline
        \multirow{4}{*}{$\pi_\theta(s)$} & network architecture & 2-256-256-2 \\
        & activation for hidden layer & $relu$ \\
        & activation for output layer & none \\
        & learning rate & $10^{-4}$ \\\hline
        \multirow{8}{*}{Agent} & $\tau$ or Polyak averaging parameter & 0.005 \\
        & $\gamma$ or discount factor & $1 - 10^{-2}$ \\
        & $\lambda$ or scaling factor for actions & $0.01$ \\
        & optimizer & Adam \\
        & replay buffer $\mathcal{R}$ capacity & $10^{4}$ \\
        & minibatch size for update & $128$ \\
        & maximum steps per episode & $500$ \\
        & number of training epochs & $10000$ \\\hline
        \multirow{3}{*}{SAC specific} & initial $\alpha$ or entropy coefficient & $0.5$ \\
        & $\alpha$ value & variable \\
        & learning rate for $\alpha$ & $10^{-4}$ \\\hline
        \multirow{3}{*}{TD3 specific} & target update delay interval & $8$ steps \\
        & actor noise standard deviation & $0.4$ \\
        & actor noise clip & $1.0$ \\\hline
    \end{tabular}
    \vspace{2mm}
    \caption{Parameters used while training the RL agent.}
    \label{tab: parameters}
\end{table}

\section{Experiments}
\label{results}

The proposed algorithm is applied to determine the pathway with the minimum energy barrier on the M{\"u}ller–Brown potential energy surface \cite{mueller_brown}. The M{\"u}ller–Brown potential has been used to benchmark the performance of several algorithms that determine the minimum energy pathways, such as the molecular growing string method \cite{mb_gsm}, Gaussian process regression for nudged elastic bands \cite{mb_neb}, and accelerated molecular dynamics \cite{mb_acc_md}. Therefore, it is also used in this work to demonstrate the applicability of the proposed method. A custom \texttt{Gym} environment \cite{towers_gymnasium_2023} was created following the gymnasium interface (inheriting from the \texttt{class Gym}) to model the problem as a Markov Decision Process to be solved by a reinforcement learning pipeline. The values for the parameters used in Algorithm \ref{used_algorithm} are listed in Table \ref{tab: parameters}.

\subsection{Results}

\begin{figure*}
    \centering
    \begin{subfigure}[b]{0.55\textwidth}
        \includegraphics[width=0.9\textwidth]{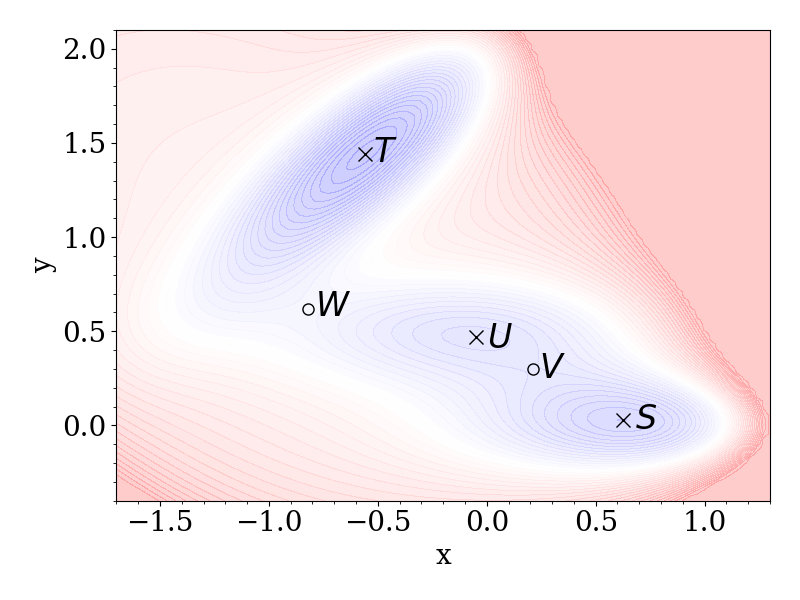}
        \caption{Potential energy surface to find the paths with minimum energy barrier.}
        \label{pes_fig}
    \end{subfigure}%
    ~ 
    \begin{subfigure}[b]{0.44\textwidth}
        \centering
        \begin{scriptsize}
            \begin{tabular}{c | c r r r}\hline
                & pt & x & y & E \\\hline
                & S & $0.623$ & $0.028$ & $-108.2$ \\
                Minima & T & $-0.558$ & $1.442$ & $-146.7$ \\
                & U & $-0.050$ & $0.467$ & $-80.8$ \\\hline
                Saddle & V & $0.210$ & $0.300$ & $-72.3$ \\
                points & W & $-0.820$ & $0.620$ & $-40.7$ \\\hline
            \end{tabular}
            \vspace{2mm}
            \caption{Minima and saddle points on the chosen potential energy surface.}
            \label{tab: pes_points}
            \begin{tabular}{c | r c}\hline
                Parameter & Value \\\hline
                number of dimensions ($d$) & $2$ \\
                limits for dimensions 1 & $(-1.70, 1.30)$ \\
                limits for dimensions 2 & $(-0.40, 2.10)$ \\
                scaling factor for action ($\lambda$) & $0.01$ \\
                tolerance for convergence ($\delta$) & $10^{-4}$ \\\hline
            \end{tabular}
        \end{scriptsize}
        \caption{Some parameters of the Markov Decision process to find pathways with the minimum energy barrier on the chosen potential.}
        \label{tab: mdp_parameters}
    \end{subfigure}
    \caption{The environment in which the agent learns to find the path with the minimum energy barrier.}
\end{figure*}

The M{\"u}ller–Brown potential is characterized by the following potential:
\begin{equation}
    V(x, y) = \sum_{i=0}^3 W_i\cdot \exp\left[a_i\left(x - \overline{x}_i\right)^2 + b_i\left(x - \overline{x}_i\right)\left(y - \overline{y}_i\right) + c_i\left(y - \overline{y}_i\right)^2 \right]
    \label{potential0}
\end{equation}

where $W = (-200, -100, -170, 15)$, $a = (-1, -1, -6.5, 0.7)$, $b = (0, 0, 11, 0.6)$, $c = (-10, -10, -6.5, 0.7)$, $\overline{x} = (1, 0, -0.5, -1)$, and $\overline{y} = (0, 0.5, 1.5, 1)$. The potential energy surface for the system is plotted in Figure \ref{pes_fig}, and the coordinates of the local minima and saddle points for the potential energy surface and their corresponding energies are tabulated in Table \ref{tab: pes_points}. The RL agent was trained to locate a path on this surface from $S\,(0.623, 0.028)$ with an initial random step (with zero mean and a standard deviation of $0.1$) taken as the starting state to $T\,(-0.558, 1.442)$ as the terminal state, with the minimum energy barrier. The first random step was chosen to avoid the same starting point in each training iteration of the agent, so it learns a more generalized policy. Some of the parameters for the Markov Decision Process to model this potential are given in Table \ref{tab: mdp_parameters}.

\begin{figure*}
    \centering
    \begin{subfigure}[t]{0.33\textwidth}
        \includegraphics[width=\textwidth]{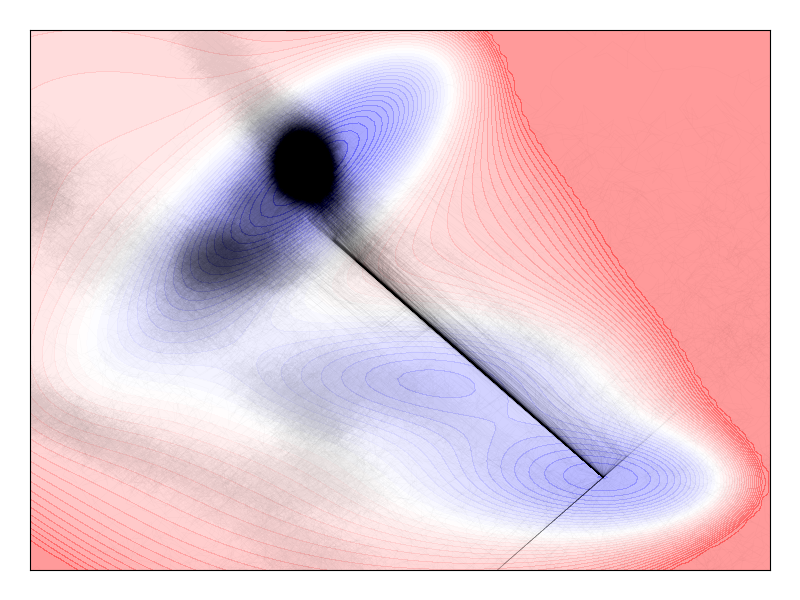}
        \caption{TD3}
        \label{td3_explore}
    \end{subfigure}%
    ~
    \begin{subfigure}[t]{0.33\textwidth}
        \includegraphics[width=\textwidth]{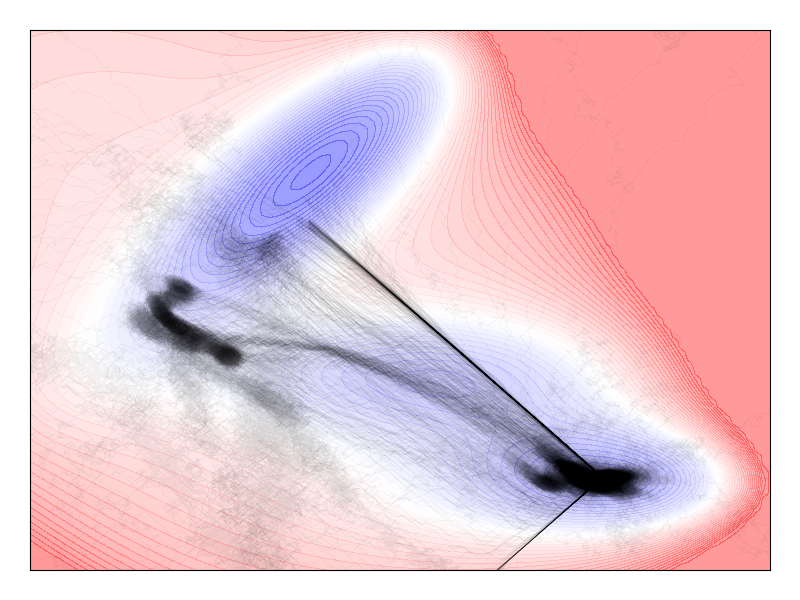}
        \caption{SAC}
        \label{sac_explore}
    \end{subfigure}%
    ~ 
    \begin{subfigure}[t]{0.33\textwidth}
        \centering
        \includegraphics[width=\textwidth]{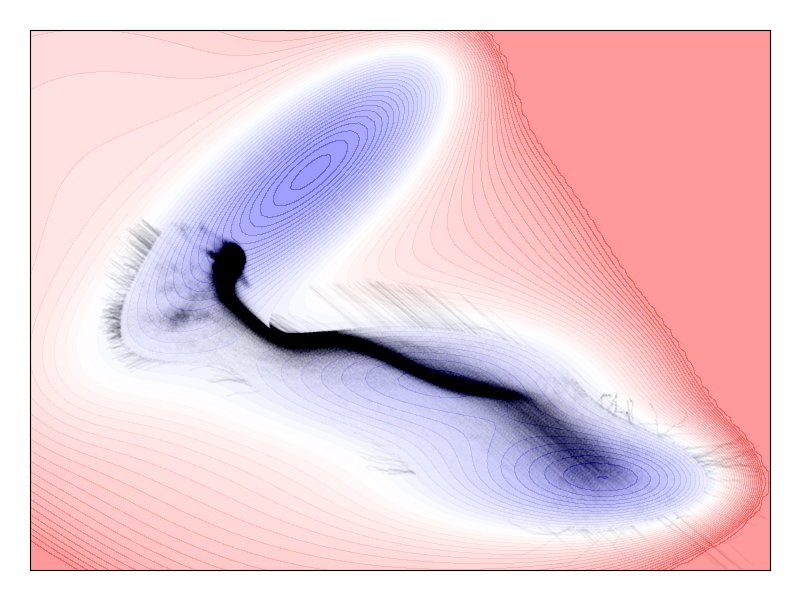}
        \caption{Algorithm \ref{used_algorithm}}
        \label{this_explore}
    \end{subfigure}
    \caption{Scatter plot of the regions visited by the reinforcement learning agent during the course of learning while using different algorithms.}
    \label{algos}
\end{figure*}

\begin{figure*}
    \centering
    \begin{subfigure}[t]{0.55\textwidth}
        \centering
        \includegraphics[width=\textwidth]{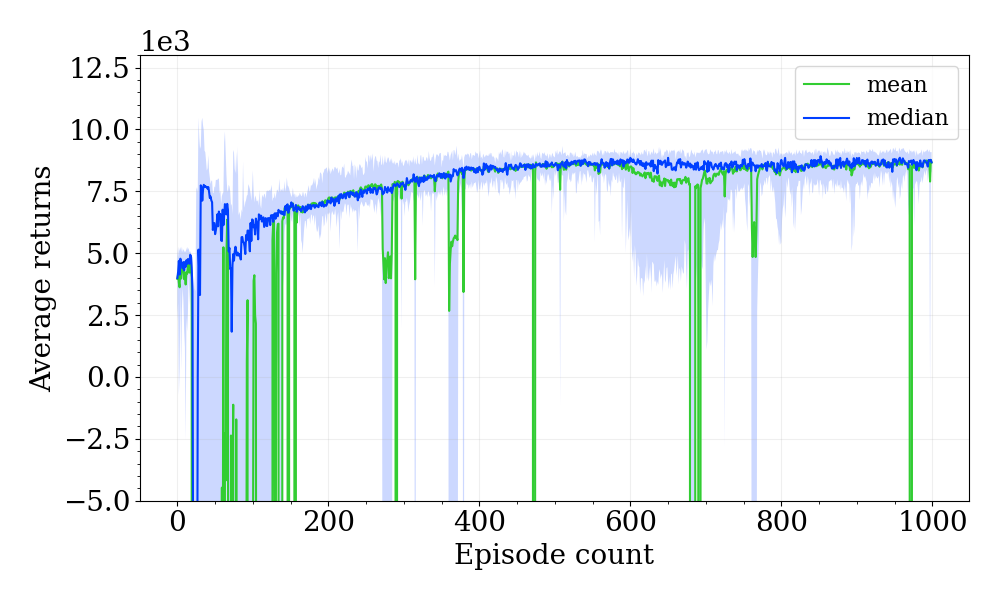}
        \caption{Learning curve for the agent averages over $11$ trials. }
        \label{rewards_fig}
    \end{subfigure}%
    ~
    \begin{subfigure}[t]{0.42\textwidth}
        \centering
        \includegraphics[width=\textwidth]{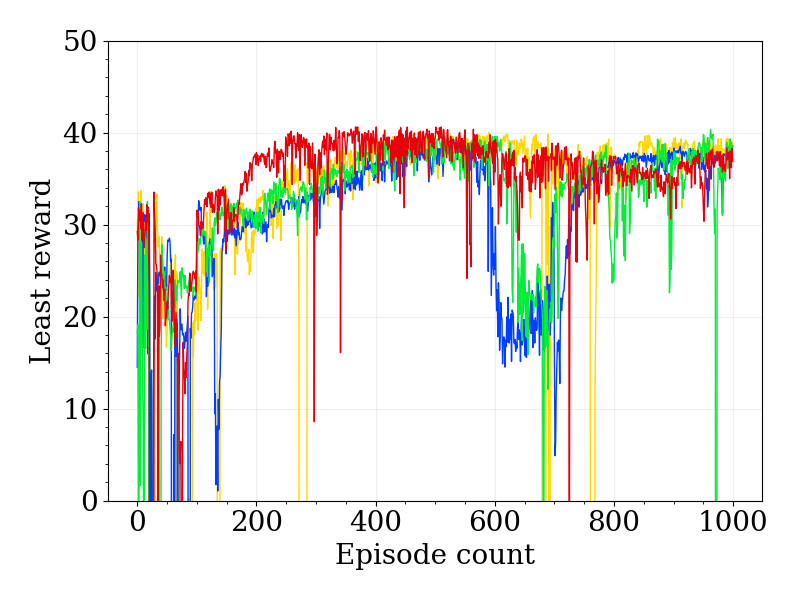}
        \caption{Least reward collected in the episode. }
        \label{lowest_reward_fig}
    \end{subfigure}
    \begin{subfigure}[t]{0.49\textwidth}
        \centering
        \includegraphics[width=\textwidth]{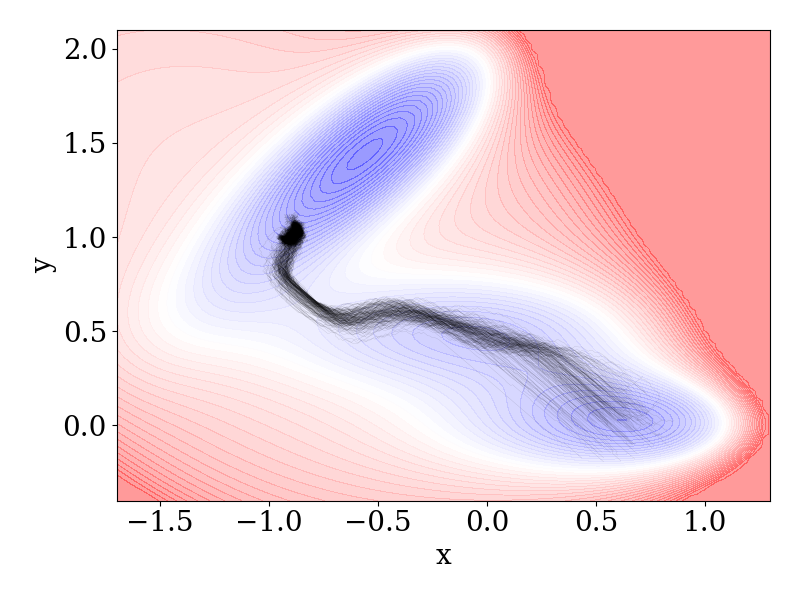}
        \caption{Trajectories generated by the agent after training.}
        \label{paths_fig}
    \end{subfigure}%
    ~ 
    \begin{subfigure}[t]{0.49\textwidth}
        \centering
        \includegraphics[width=\textwidth]{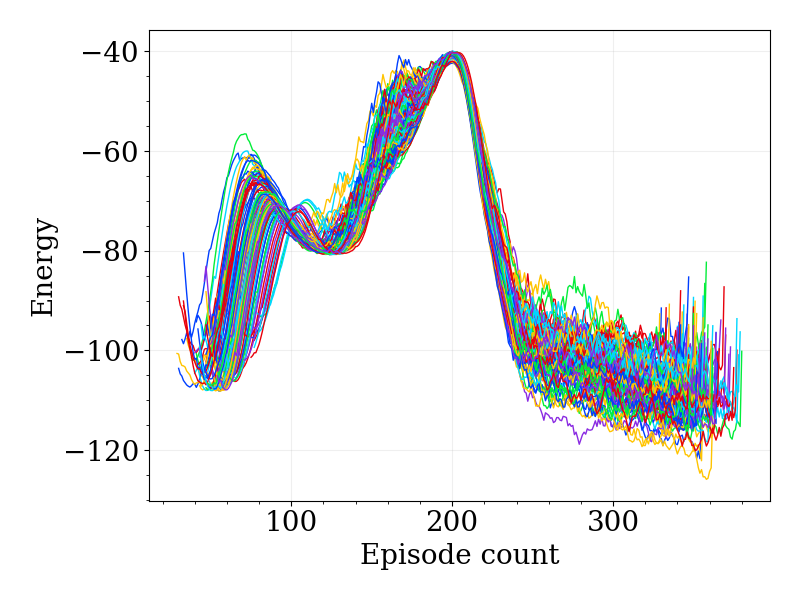}
        \caption{Energy profiles for the generated trajectories.}
        \label{profiles_fig}
    \end{subfigure}
    \caption{(a) The learning curve for the agent in the reinforcement learning environment. (b) The plot of the variation of the least reward collect by the agent in a step with the validation episode count. (c) Trajectories generated by the trained agent following the learnt policy along with the corresponding energy profiles (d).}
\end{figure*}

\subsubsection{Comparison of different algorithms}

Figure \ref{algos} shows scatter plots of the trajectories generated by various reinforcement learning algorithms: TD3 in Figure \ref{td3_explore}, SAC in Figure \ref{sac_explore}, and the proposed modified SAC algorithm in Figure \ref{this_explore}. While the agent trained by the TD3 algorithm does reach the intended target state, it starts to exploit a flaw in the formulation of the MDP by trying to reach the vicinity of the final state quickly and staying near enough to it so that it collects rewards, but does not terminate the episode. This results in a high density in the plot along the straight line connecting the initial and final states and around the final state. It gives a much higher estimate than the correct minimum energy barrier for the transition. The agent trained using SAC shows improved performance, possibly due to the entropy regularizer forcing it to learn a more diverse policy (rather than one that would result in a straight line connecting the initial and final states). However, while generating trajectories in the testing environment, most of the trajectories did not leave the local minima in the vicinity of the start state. Moreover, the learned policy has high variance. The proposed Algorithm \ref{used_algorithm} learns a much more stable policy and confines itself to exploring the region with lower energies leading to the terminal state (specifically the vicinity of the saddle point) rather than the entire environment. It explores sufficiently and then exploits the state-action values learned appropriately, providing better estimates of the energy barrier for the transition.

The learning curve for the agent under Algorithm \ref{used_algorithm} is shown in Figure \ref{rewards_fig}. The data for this curve were generated by allowing the agent to solve the MDP in evaluation mode once every $10$ training episodes, where the neural networks were not updated to monitor the agent's learning. The ascending learning curve indicates that the agent gradually learns to find a path to the terminal state that maximizes the rewards. The blue line represents the median reward, while the green line shows the mean reward over $11$ training iterations, each consisting of $10\times 1000$ training episodes. The light blue shaded region denotes the spread of the rewards (maximum and minimum). A low spread in rewards indicates consistent performance by the reinforcement learning agent in the validation episodes.

\subsubsection{Performance of the trained agent}

In Figure \ref{paths_fig}, an ensemble of paths generated by the trained RL agent is plotted on the surface of the potential energy, with the starting points slightly perturbed from $(0.623, 0.028)$ by noise added from $\mathcal{N}(0, 0.1)$ on the surface of the potential energy. The model used to generate these trajectories was the model at the $500$th validation step (and not after the entire training consisting of $1000$ validation steps) for reasons elaborated later. It can be seen from Figure \ref{paths_fig} that the agent spends more time near the terminal state rather than reaching the terminal state to receive an immediate reward, as it allows the agent to collect rewards for more steps. In the case of a coarse-grained maze representation of the potential energy surface, this problem was solved by rewarding the agent only the first time it visited a grid cell. Using a similar idea of not rewarding the agent when it is in the close neighborhood of an already visited state artificially perturbs the reward map and did not work in this case. The best performance was achieved by gradually varying the maximum number of steps the agent was allowed to take in an episode. If the number of steps allowed in an episode is too short, the agent does not escape the local minima to reach the terminal state. If the number of steps is too large, then the agent reaches the terminal state and discovers that it receives larger rewards by remaining in its vicinity, but not so close that the episode is terminated. In an attempt to reach the terminal state earlier, the agent tries to approach the terminal state sooner, choosing a more direct pathway, which lifts the trajectory out of the saddle point. It was observed that a maximum of $500$ steps per episode led to the best performance by the agent. The agent did not leave the local minima if fewer steps were allowed ($200$ steps), and the agent passed through states with much higher energies than optimal to reach a minimum energy state if longer episodes were allowed ($1000$ steps).

Early stopping during the training of a neural network has often been found to be helpful in scenarios where continued training worsens the performance of the model \cite{early_stop1, early_stop2}. Borrowing the idea of early stopping, the agent's training was stopped when the minimum reward collected by the agent (corresponding to the maximum energy along the pathway) started increasing again. This behavior was observed in the case of the agent, as plotted in Figure \ref{lowest_reward_fig}. The minimum reward collected by the agent during the episode increases initially (indicating that the agent finds a pathway with a progressively better energy barrier) until the $500$th validation step before decreasing slightly. Plots for only $4$ of the $11$ trials are shown for clarity. This might indicate that the agent does not improve its performance after that step. Furthermore, the learning curve in Figure \ref{rewards_fig} shows an increase in spread after $500$ iterations. These reasons led to using the model after $500$ iterations to generate the final trajectory in test mode to estimate the energy barrier for the reaction.
The energy profiles along the generated trajectories are plotted in Figure \ref{profiles_fig} aligned by the maximum of the profiles (and not by the start of the trajectories) for better visualization. The energy barrier predicted for the transition of interest is $-40.36 \pm 0.21$. One can see that the agent learns to predict the path with the correct minimum energy barrier, albeit the energy barrier estimated by the agent is a little higher than the optimal analytical solution ($-40.665$). However, the result demonstrates that reinforcement learning algorithms can be used to locate the minimum energy barrier for transitions between stable states in complex systems. The paths suggested by the trained agent cluster around the minimum energy path and pass through the vicinity of the actual saddle point that represents the energy barrier. However, there still seems to be some way to go to improve the sampling densities around the saddle point, which determines the barrier height, to avoid overestimating it.

\subsubsection{On the choice of the scaling factor}
The scaling factor, $\lambda$, scales the action for the agent. In most cases, it is used to adjust the step size for the agent while keeping the action space within some standard interval ($[0, 1]$ or $[-1, 1]$). This scaling factor was varied along with the number of steps in an episode, and the combination of $0.01$ for the $\lambda$ and $500$ steps in an episode led to the best performance of the agent. With these parameters, the agent reaches near the terminal state with just the number of required small enough steps to end the episode. A larger value of $\lambda$ resulted in the agent taking longer steps over regions of the potential energy surface with a higher energy to give an incorrect estimate of the barrier height. Smaller values of $\lambda$ led to smaller steps, and the agent did not leave the local minima to explore other regions of the potential energy surface, and is unsuccessful a its assigned task.

As noted at the end of Section \ref{mdp}, the formulated Markov decision process suffers from the drawback that that agent might stay at a small distance $\epsilon (> \delta)$ from that target state, and collect rewards until the remainder of the episode. To discourage the agent from doing this, the episode is truncated after $500$ steps. Increasing the number of steps in the episode would encourage the agent to stay a small distance away from the target state, rather than reaching the target state and terminating the episode, once it realizes that it can increase the total reward collected by this. On the other hand, if the number of steps in an episode is too small, it would not reach near the target state. 
The choice of the scaling factor $\lambda$ is related to the number of steps in an episode. Together, they determine the maximum distance from the start state the agent can reach. If both $\lambda$ and the number of steps in an episode is decreased below a limit, the agent would never be able to reach the target state \footnote{For example, if the maximum number of steps for the maze in Figure \ref{pes_rl} is limited to $5$, then the agent can never reach the terminal state. As an analogy, decreasing the cell size, resulting in increasing the number of cells in the maze, would be equivalent to decreasing $\lambda$ in the current case.}. With an appropriate choice of $\lambda$ and the length of the episode, the agent does not make too many long jumps through higher energy states to reach a state with lower energy faster. Decreasing $\lambda$ would require increasing the maximum number of steps in an episode, so that the agent explores regions away from the starting point, but not too much so that the trajectory passes through regions with higher energy. A few experiments were done to determine the pair of the values for $\lambda$ and the number of steps in an episode which gives the best performance by the agent, and the results are plotted in Appendix \ref{appendix1}.

\subsection{Ablation Studies}
\begin{figure*}
    \centering
    \begin{subfigure}[b]{0.5\textwidth}
        \centering
        \begin{scriptsize}
            \begin{tabular}{r r | r}\hline
                Modification & Value & Returns \\\hline
                target policy & absent & $8082\pm\,\,\,10$ \\
                smoothing & \textbf{present} & $\mathbf{8651\pm 107}$\vspace{0.1cm}\\\hline
                & $0$ & $4826\pm 984$ \\
                & $2$ & $7738\pm 234$ \\
                policy update & $4$ & $7741\pm 220$ \\
                delay & $\mathbf{8}$ & $\mathbf{7994\pm206}$\\
                & $16$ & $7677\pm 204$ \\
                & $32$ & $5783\pm 142$\vspace{0.1cm}\\\hline
                \multirow{6}{*}{$\alpha$ tuning}
                & $10^{-3}$ & $3877\pm \,\,\,29$ \\
                & $10^{-2}$ & $4153\pm \,\,\,14$ \\
                & $10^{-1}$ & $7425\pm 414$ \\
                & $2\times 10^{-1}$ & $4880\pm 654$ \\
                & $5\times 10^{-1}$ & $6247\pm 190$ \\
                & \textbf{tunable} & $\mathbf{8651\pm\,107}$\\\hline
            \end{tabular}
        \end{scriptsize}
        \caption{Returns for the last $100$ episodes of trials\\ for each modification on the learning of the agent.}
        \label{components_tab}
    \end{subfigure}%
    ~
    \begin{subfigure}[b]{0.5\textwidth}
        \centering
        \includegraphics[width=\textwidth]{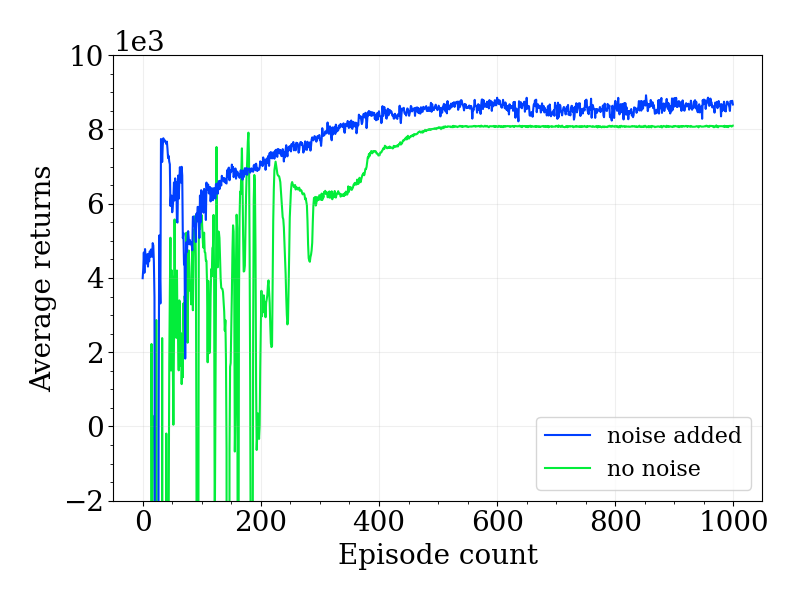}
        \caption{Effect of adding noise to the target action \\on the learning of the agent.}
        \label{noise_fig}
    \end{subfigure}
    \begin{subfigure}[t]{0.5\textwidth}
        \includegraphics[width=\textwidth]{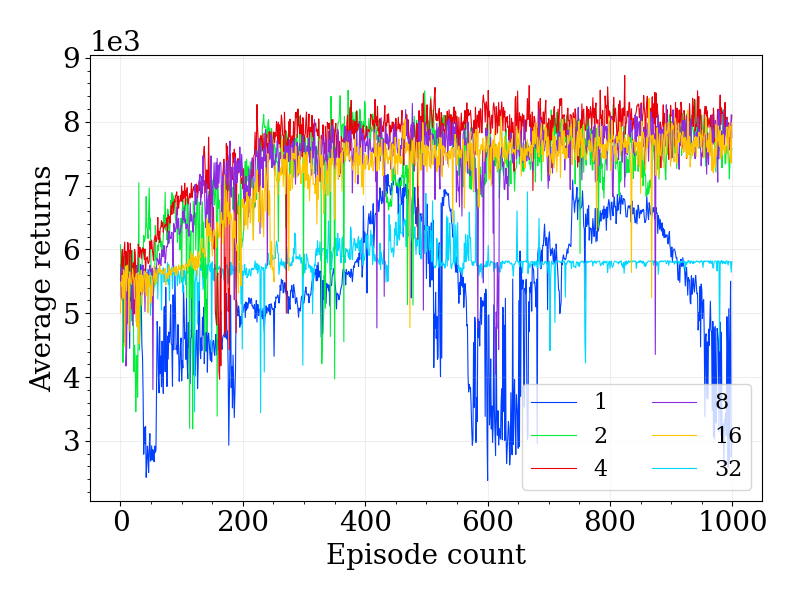}
        \caption{Effect of delaying the update of the actor net, \\and target critic-nets on the learning of the agent.}
        \label{delay_fig}
    \end{subfigure}%
    ~ 
    \begin{subfigure}[t]{0.5\textwidth}
        \centering
        \includegraphics[width=\textwidth]{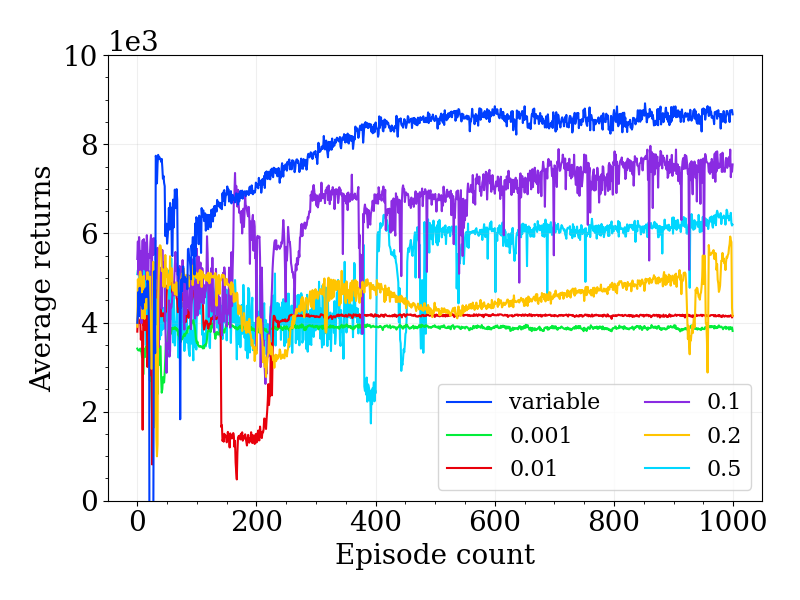}
        \caption{Effect of varying constant values of $\alpha$ and \\a tunable $\alpha$ on the learning of the agent.}
        \label{alpha_fig}
    \end{subfigure}
    \caption{Effect of the various modifications to the SAC algorithm on the learning of the agent.}
\end{figure*}

Several modifications were made to the standard SAC algorithm to be used in this particular case (highlighted in blue in Algorithm \ref{used_algorithm}). Studies were performed to understand the contribution of each individual component to the working of the algorithm in this particular environment by comparing the performance of the algorithm with different hyperparameters for a component. The parameters for one modification were varied, keeping the parameters for the other two modifications unchanged from the fine-tuned algorithm. Each modification and its contribution to the overall learning of the agent are described in the following sections. The mean and the standard deviation of the returns from the last $100$ training steps for each modification of the existing algorithm are listed in Table \ref{components_tab} to compare the performance of the agents. The modification that leads to the highest returns is highlighted.

\subsubsection{Target Policy Smoothing}
Injecting random noise (with a standard deviation $\sigma$) into the action used in the environment (in line 9 of Algorithm \ref{used_algorithm}) encourages the agent to explore, while adding noise to the actions used to calculate the targets (in line 14 of Algorithm \ref{used_algorithm}) acts as a regularizer, forcing the agent to generalize over similar actions. In the early stages of training, the critic Q-nets can assign inaccurate values to some state-action pairs, and the addition of noise prevents the actor from rote learning these actions based on incorrect feedback. On the other hand, to avoid the actor taking a too random action, the action is clipped by some maximum value for the noise (as done in lines 9 and 14 of Algorithm \ref{used_algorithm}). The effect of adding noise to spread the state-action value over a range of actions is plotted in Figure \ref{noise_fig}. Adding noise leads to the agent learning a policy with less variance in the early learning stages and a more consistent performance.

\subsubsection{Delayed Policy Updates}
Delaying the updates for the actor nets and the target Q-nets (in lines 17 and 19 of Algorithm \ref{used_algorithm}) allows the critic Q-nets to update more frequently and learn at a faster rate, so that they can provide a reasonable estimate of the value for a state-action pair before it is used to guide the policy learned by the actor net. The parameters of the critic Q-nets might often change abruptly early on while learning, undoing whatever the agent had learned (catastrophic failure). Therefore, delayed updates of the actor net allow it to use more stable state-action values from the critic nets to guide the policy learned by it. The effect of varying intervals of delay for the actor update on the learning of the agent is plotted in Figure \ref{delay_fig}. Updating the actor net for every update of the critic nets led to a policy with a high variance (blue plot). Delaying the update of the actor net to once every 2 updates of the critic resulted in the agent learning a policy that provided higher returns but still had a high variance (green plot). Delaying the update of the actor further (once every 4 and 8 updates of the critic net plotted as the red and magenta curves, respectively) further improved the performance of the agent. One can notice the lower variance in the policy of the agent during the early stages (first 200 episodes of the magenta curve) for the agent which updates the actor net and target critic nets once every 8 updates of the critic nets. However, delaying the updates for too long intervals would cripple the learning of the actor. The performance of the agent suffers when the update of the actor is delayed to once every 16 updates of the critic nets (yellow curve) and the agent fails to learn when the update of the actor net is further delayed to once every 32 updates of the critic nets (cyan curve).

\subsubsection{Tuning the Entropy Coefficient}
The entropy coefficient $\alpha$ can be tuned as the agent learns (as done in line 18 of Algorithm \ref{used_algorithm}), which overcomes the problem of finding the optimal value for the hyperparameter $\alpha$ \cite{sac2}. Moreover, simply fixing $\alpha$ to a single value might lead to a poor solution because the agent learns a policy over time: it should still explore regions where it has not learned the optimal action, but the policy should not change much in regions already explored by the agent that have higher returns. In Figure \ref{alpha_fig}, the effect of the variation of the hyperparameter $\alpha$ on the learning of the agent is compared. As can be seen, a tunable $\alpha$ allows the agent to learn steadily, encouraging it to explore more in the earlier episodes and exploiting the returns from these explored regions in the latter episodes, resulting in a more stable learning curve (blue curve). A too low value of $\alpha$, such as $10^{-3}$ or $10^{-2}$, makes the algorithm more deterministic (TD3-like), which leads to suboptimal performance and the agent being stuck in a local minimum (plotted as green and red curves, respectively). An $\alpha$ value of $0.1$ has a performance comparable to the tunable $\alpha$, but the learning curve is less stable and there are abrupt changes in the policy function (magenta curve). The original implementation of SAC suggested $0.2$ as a fixed value for $\alpha$, which leads to a learning curve that results in a policy with high variance (yellow curve). A too high value of $\alpha$, such as $0.5$, makes the algorithm more stochastic (REINFORCE-like), which also leads to suboptimal learning (cyan curve).

\subsection{Some more surfaces}
\begin{figure*}
    \centering
    \begin{subfigure}[t]{0.33\textwidth}
        \includegraphics[width=\textwidth]{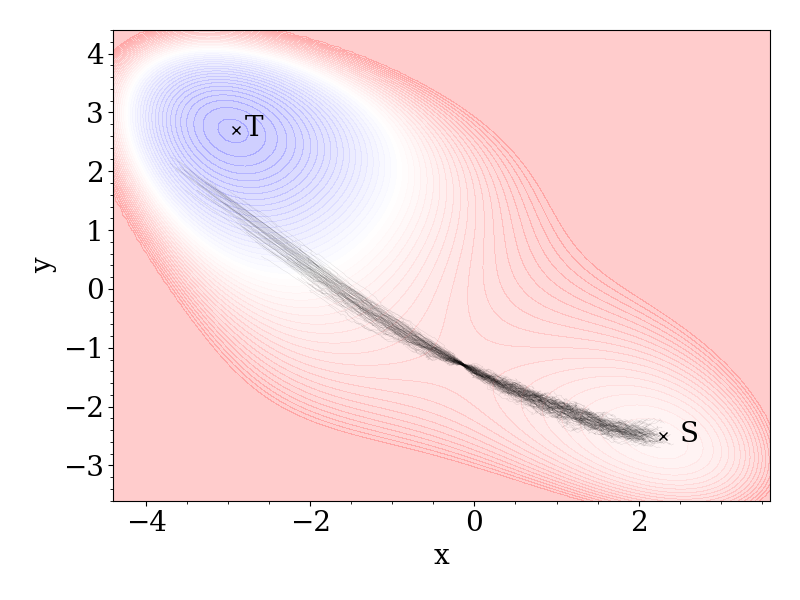}
        \label{pes1}
    \end{subfigure}%
    ~
    \begin{subfigure}[t]{0.33\textwidth}
        \includegraphics[width=\textwidth]{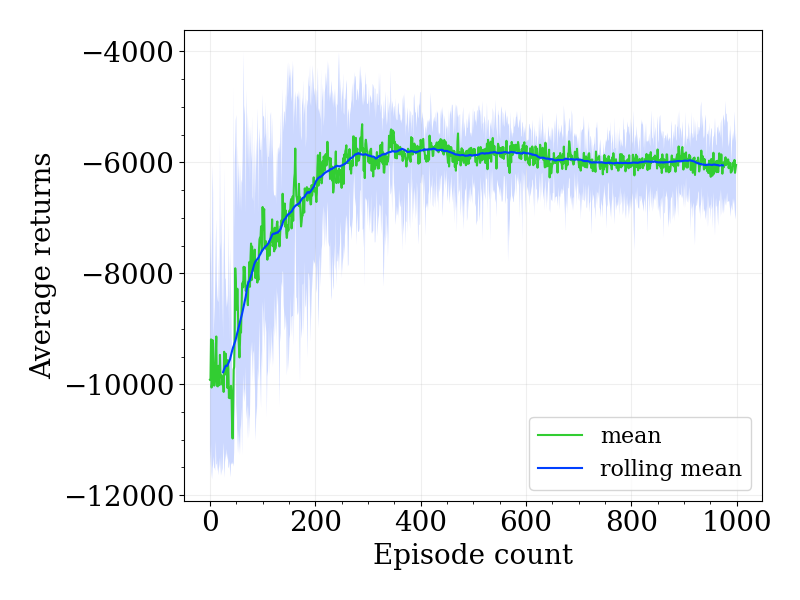}
        \caption{Equation \ref{potential1}}
        \label{pes1_traj}
    \end{subfigure}%
    ~ 
    \begin{subfigure}[t]{0.33\textwidth}
        \centering
        \includegraphics[width=\textwidth]{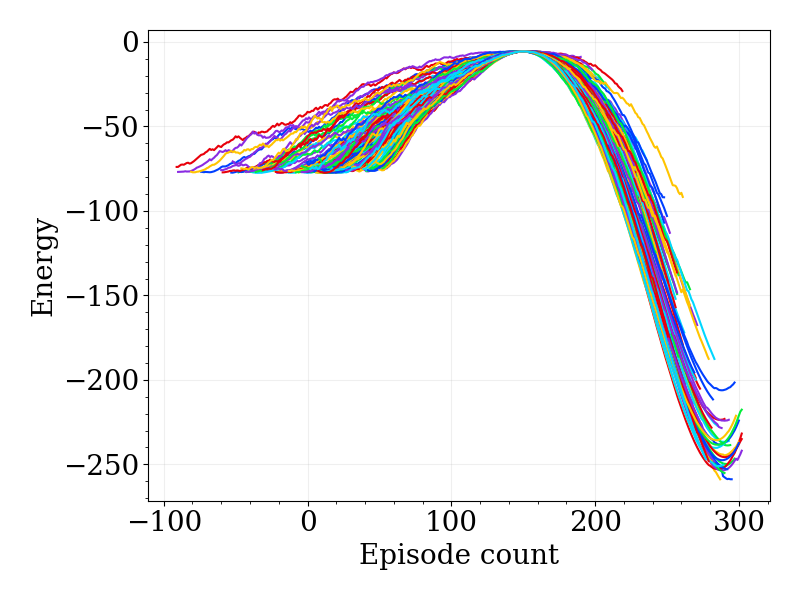}
        \label{pes1_profile}
    \end{subfigure}
    \begin{subfigure}[t]{0.33\textwidth}
        \includegraphics[width=\textwidth]{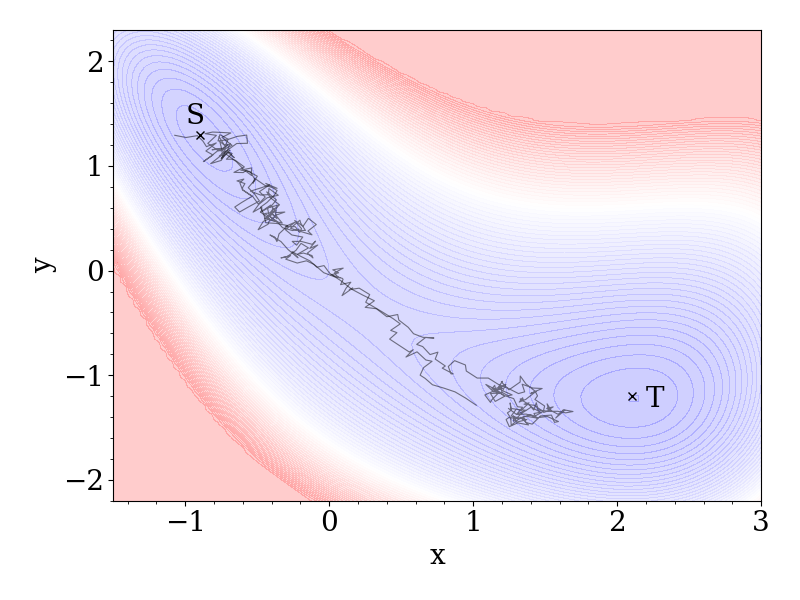}
        \label{pes2}
    \end{subfigure}%
    ~
    \begin{subfigure}[t]{0.33\textwidth}
        \includegraphics[width=\textwidth]{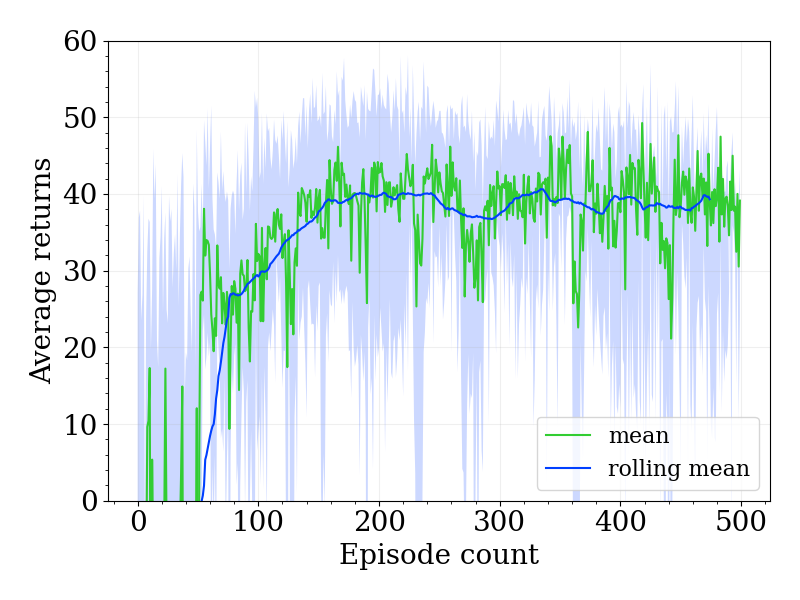}
        \caption{Equation \ref{potential2}}
        \label{pes2_traj}
    \end{subfigure}%
    ~ 
    \begin{subfigure}[t]{0.33\textwidth}
        \centering
        \includegraphics[width=\textwidth]{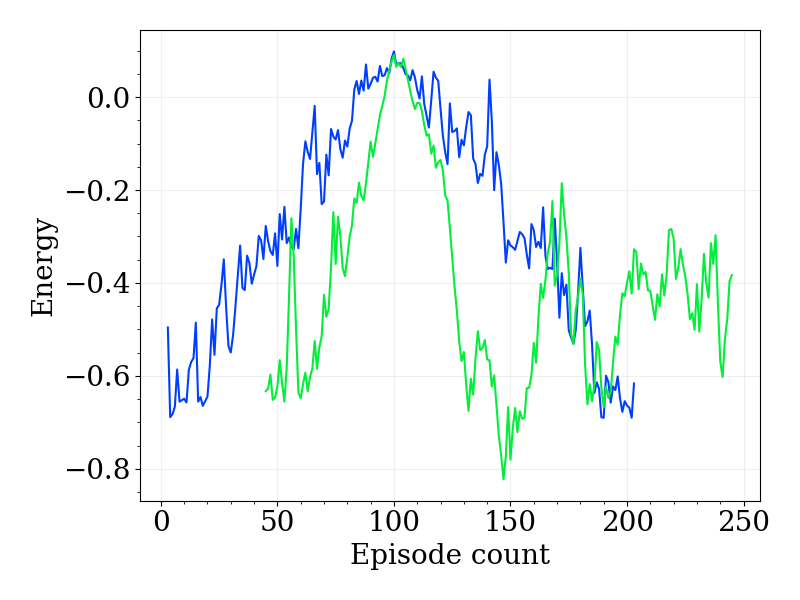}
        \label{pes2_profile}
    \end{subfigure}
    \caption{Results on using the proposed algorithm on the potential energy surfaces depicted by (a) Equation \ref{potential1} and (b) Equation \ref{potential2}.}
    \label{results_more}
\end{figure*}

Here we demonstrate the results by an agent trained by Algorithm \ref{used_algorithm} on some more two-dimensional potential energy surfaces with two potential wells.
\begin{align}
    V(x, y) &= (x - 1)^4 + (y - 1)^4 + (x + 1)^4 + (y + 1)^4 -20(x - y)^2 + 60x \label{potential1}\\
    V(x, y) &= \left[y - 0.4(x^2 - 4x)\right]^2 - x^2 + 0.1(x^4 + y^4) + 0.5x \label{potential2}\\
    V(x, y) &= \left[(1 - x^2 - y^2)^2 + \frac{y^2}{x^2 + y^2} \right]\left(1+\frac{1}{1 + e^{-y}}\right) \label{potential3}
\end{align}
While the potential energy surfaces represented by Equation \ref{potential3} was taken from \cite{toy_potential}, those represented by Equation \ref{potential1} and Equation \ref{potential2} were crafted by hand. The results of using the proposed Algorithm \ref{used_algorithm} on these potential energy surfaces are depicted in Figure \ref{results_more}. The left column shows the trajectories generated by the agent after training between the two minima on the potential energy surface. The central column plots the learning curves of the agent, with the mean cumulative sum of returns plotted as a green line. The spread of the sum of the returns is shaded in blue, while the rolling mean is represented as a solid blue line. The right column shows the energy profiles of the generated trajectories. The estimated energy barrier for the transition on the potential energy surface given by Equation \ref{potential1} was $-5.575$ while that for the potential energy surface given by Equation \ref{potential2} was $0.94$.

\section{Discussions}
\label{discussion}
(Horizon Europe Research and Innovation Project CORENET)
Previous work in determining transition pathways using deep learning or reinforcement learning techniques includes formulating the problem as a shooting game solved using deep reinforcement learning \cite{rl_shooting_game}. The authors in \cite{rl_shooting_game} sample higher energy configurations and shoot trajectories with randomized initial momenta in opposite directions, expecting them to converge at the two desired local minima. In contrast, the method proposed here starts from a minimum on the potential energy surface and attempts to generate a trajectory to another minimum. Additionally, in \cite{stoc_opt_control}, the problem is formulated as a stochastic optimal control problem, where neural network policies learn a controlled and optimized stochastic process to sample the transition pathway using machine learning techniques. Stochastic diffusion models have also been used to model elementary reactions and generate the structure of the transition state, preserving the required physical symmetries in the process \cite{OA-ReactDiff}. Furthermore, the problem of finding transition pathways was recast into a finite-time horizon control optimization problem using the variational principle and solved using reinforcement learning in \cite{deep_rl_finite_horizon}. Moreover, a hybrid-DDPG algorithm was implemented in \cite{ddpg_mb} to identify the global minimum on the M{\"u}ller–Brown potential, but did not identify pathways between minima as in this work. Recent work \cite{rl_transition_state} used an actor-critic reinforcement learning framework to optimize molecular structures and calculate minimum energy pathways for two reactions.

There has also been previous work \cite{react_opt} to optimize chemical reactions by perturbing the experimental conditions to achieve better selectivity, purity, or cost for the reaction using deep reinforcement learning. While this approach has macroscopic applications in laboratory settings, the method proposed here focuses on a much narrower problem: given a potential energy surface, how well can the minimum energy barrier be estimated for a transition between two minima? Deep reinforcement learning has also been used to find a minimum energy pathway consisting of multiple elementary transitions in catalytic reaction networks \cite{cat_react_opt}. While the free energy barrier for a transition (which is mapped to a reward) between two states is calculated using density functional theory (DFT) with VASP software in \cite{cat_react_opt, cat_react_opt2}, the objective of the proposed method in this work is to estimate that free energy barrier using an agent trained via deep reinforcement learning, which does not require quantum mechanical calculations. In addition, reinforcement learning techniques are implemented in \cite{opt_react_routes} to minimize the cost of synthesis pathways (consisting of multiple elementary transitions) considering the price of the starting molecules and the atom economy of individual transitions. Furthermore, a reinforcement learning approach is used to search for process routes that optimize economic profit for a Markov decision process that models the thermodynamic state space as a graph.

\section{Conclusion}
\label{conclusion}

\begin{figure}
    \centering
    \begin{subfigure}[t]{0.49\linewidth}
        \includegraphics[width=\textwidth]{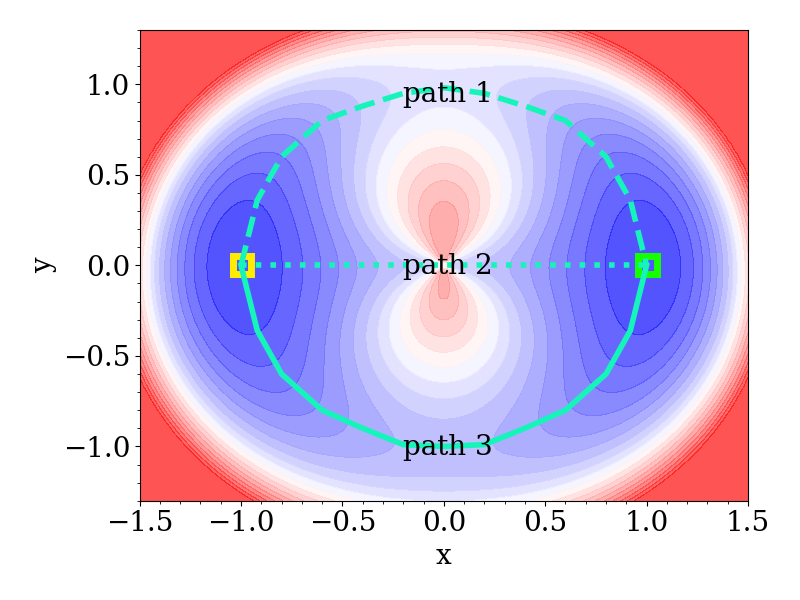}
        \caption{A potential energy surface with pathways passing through three different saddle points.}
        \label{pes_anon_fig}
    \end{subfigure}%
    ~ 
    \begin{subfigure}[t]{0.49\linewidth}
        \centering
        \includegraphics[width=\textwidth]{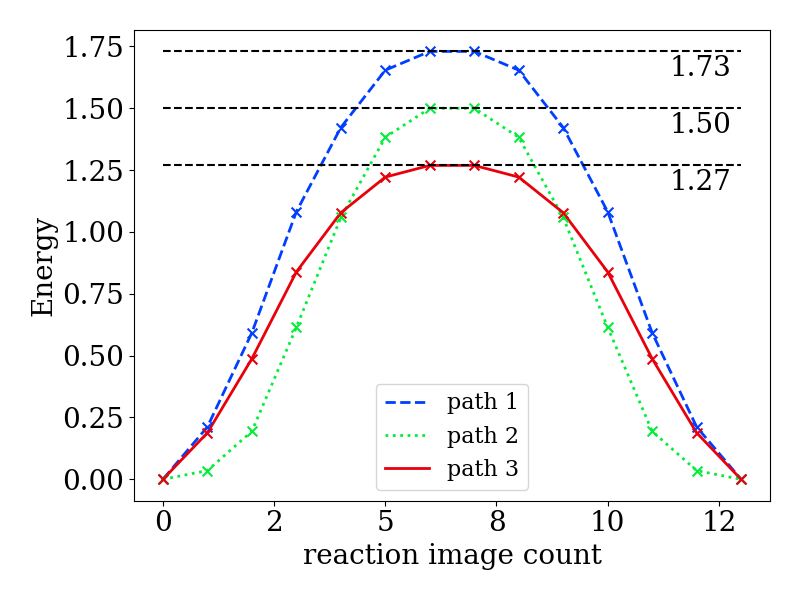}
        \caption{The energy profiles of the three pathways depicted in (a) with their respective barrier heights. }
        \label{profile_anon_fig}
    \end{subfigure}
    \caption{(a) Three possible pathways on a potential energy surface given by Equation \ref{potential3} passing though different saddle points and (b) the energy profile for the three possible transition pathways.}
    \label{anomaly_fig}
\end{figure}

Advancements in reinforcement learning algorithms based on the state-action value function have led to their application in diverse sequential control tasks such as Atari games, autonomous driving, robot movement control, and more physical domains \cite{ml_chem, rl_chem, rl_env_chem, rl_thermodynamics}. This project formulated the problem of finding the minimum energy barrier for a transition between two local minima as a cost minimization problem, solved using a reinforcement learning setup with neural networks as function approximators for the actor and critics. A stochastic policy was employed to facilitate the exploration by the agent, further perturbed by random noise. Target networks, delayed updates of the actor, and a replay buffer were used to stabilize the learning process for the reinforcement learning agent. While the proposed framework samples the region around the saddle point sufficiently, providing a good estimate of the energy barrier for the transition, there is definitely scope for improvement.
The method has been applied only to a two-dimensional system, but as a future work, it could be extended to more realistic and higher-dimensional systems. One promising alternative would be to use max-reward reinforcement learning \cite{max_reward_rl}, as it aligns well with the objective of maximizing the minimum reward obtained in an episode. However, a drawback of this method is that the reinforcement learning agent must be trained from scratch if one needs to find minimum energy pathways on a different potential energy surface. In other words, an agent trained on one potential energy surface cannot be used to determine the minimum energy barrier on a different surface, similar to how an agent trained in one \texttt{Gymnasium} environment cannot solve tasks in another. Another limitation is that the agent can only locate pathways to minima lower than the starting minimum; otherwise, remaining at the starting position minimum would yield higher rewards for the agent.

This work differs from previous work that uses reinforcement learning \cite{rl_shooting_game, rl_transition_state, deep_rl_finite_horizon, stoc_opt_control} by providing a much simpler formulation of the problem, using the energy of the state directly as the reward while searching for transition pathways with the minimum energy barrier. One of the main advantages of using a reinforcement learning based method is that, unlike traditional methods such as the nudged elastic band or the growing string method, it does not require an initial guess for the trajectory. Traditional methods use energy gradient information along the trajectory to iteratively improve to a trajectory with better energetics. However, the success of these methods depends on the initial guess for the trajectory, and gradient-based methods might get stuck in a local minimum. As shown on the potential energy surface represented by Equation \ref{potential3} in Figure \ref{anomaly_fig}, there may be multiple saddle points between two minima. The trajectory to which a nudged elastic band or a growing string method converges depends on the initial guess of the starting trajectory. Typically, the initial guess trajectory is a simple linear interpolation between the starting and ending points, which leads to the dotted trajectory (path $2$ with a barrier of $1.50$ units). Traditional gradient-based methods report this trajectory as the optimal one because the local gradients along the trajectory are minimal and cannot be improved by perturbation. However, the reinforcement learning-based method proposed in this work identifies the trajectory represented by a solid line (Path $3$ with a barrier of $1.27$ units) as the minimum energy pathway. The suboptimal solution overestimates the energy barrier for the transition by $(150-127)/127$ or $18\%$, and therefore underestimates the frequency with which it occurs by $1-e^{-1.50-(-1.27)} = 20\%$. Underestimates of the probability for a transition to occur would lead to imperfect modeling of the dynamics of the system. The use of a stochastic policy in a reinforcement learning setup avoids this problem, increasing the chances of finding a better estimate of the transition barrier as the agent explores the state space. However, as a trade-off for the simple model and generic approach, the agent learns slowly, requires a large number of environment interactions, and would have to be retrained to work in a new potential energy surface.

\begin{ack}
The author would like to gratefully acknowledge the computational resources provided by Institut for Matematik og Datalogi, Syddansk Universitet for this work. The work was also supported by the European Union (Horizon Europe Research and Innovation Project CORENET) and the Swiss State Secretariat for Education, Research and Innovation (SERI) under contract numbers 22.00017 and 22.00034, respectively. The author also thanks the two reviewers for their insightful suggestions to improve the manuscript.
\end{ack}

\begin{appendix}
\section{Additional heatmap plots}
\label{appendix1}
\begin{figure*}[h!]
    \centering
    \begin{subfigure}[t]{0.33\textwidth}
        \includegraphics[width=\textwidth]{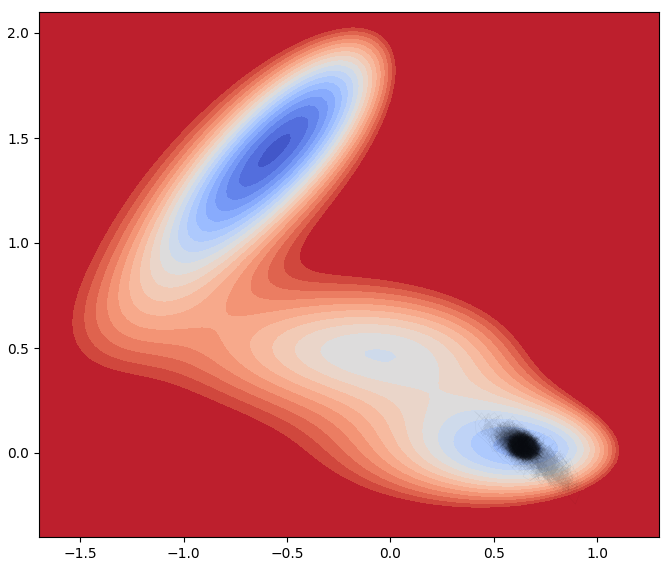}
        \caption{$\lambda = 0.005, n = 500$}
    \end{subfigure}%
    ~
    \begin{subfigure}[t]{0.33\textwidth}
        \includegraphics[width=\textwidth]{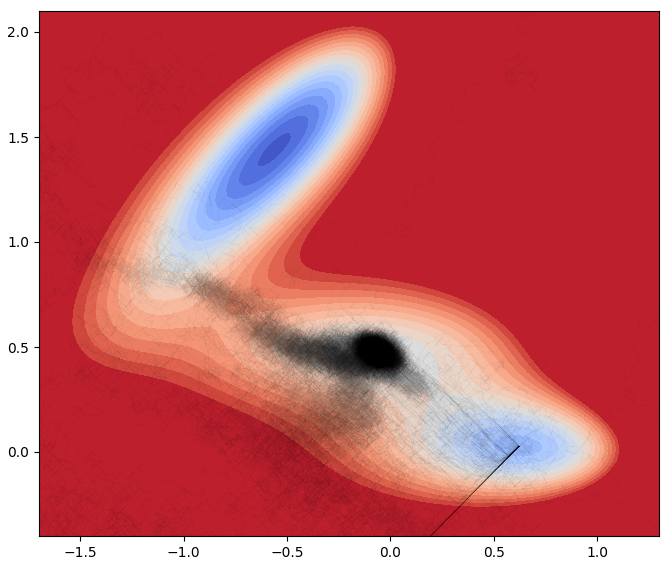}
        \caption{$\lambda = 0.005, n = 1000$}
    \end{subfigure}%
    ~ 
    \begin{subfigure}[t]{0.33\textwidth}
        \centering
        \includegraphics[width=\textwidth]{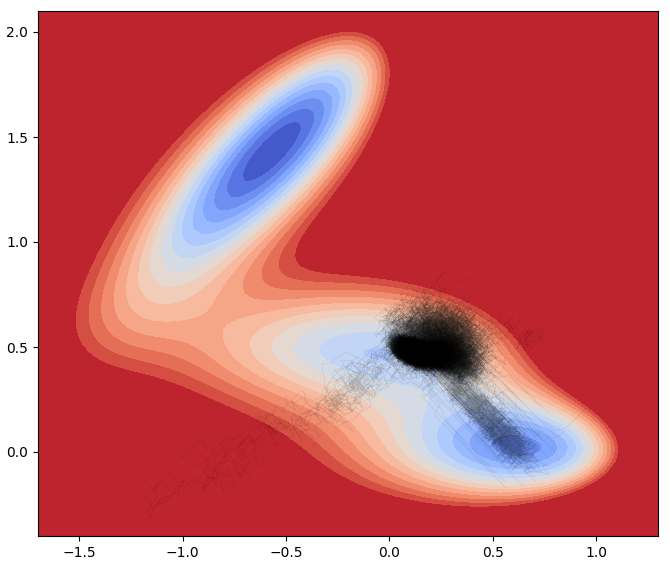}
        \caption{$\lambda = 0.005, n = 100$}
    \end{subfigure}
    \begin{subfigure}[t]{0.33\textwidth}
        \includegraphics[width=\textwidth]{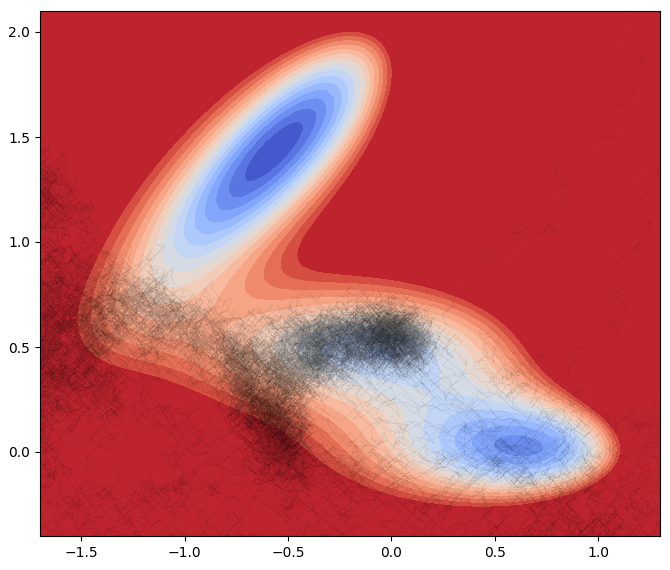}
        \caption{$\lambda = 0.001, n = 200$}
    \end{subfigure}%
    ~
    \begin{subfigure}[t]{0.33\textwidth}
        \includegraphics[width=\textwidth]{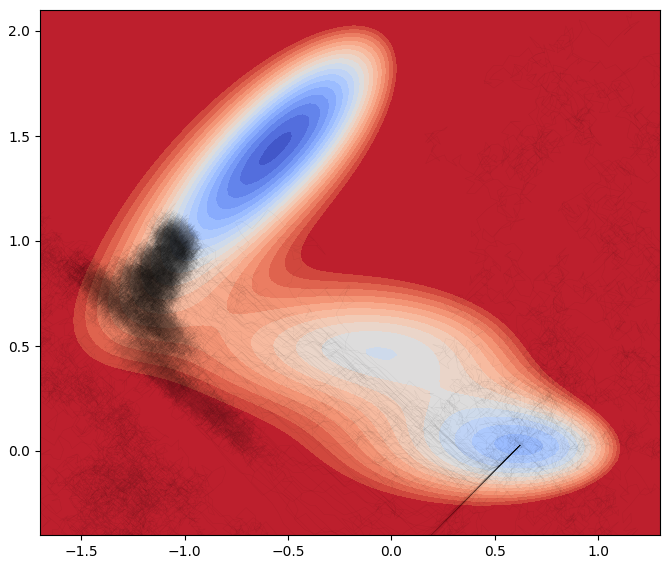}
        \caption{$\lambda = 0.02, n = 100$}
    \end{subfigure}%
    ~ 
    \begin{subfigure}[t]{0.33\textwidth}
        \centering
        \includegraphics[width=\textwidth]{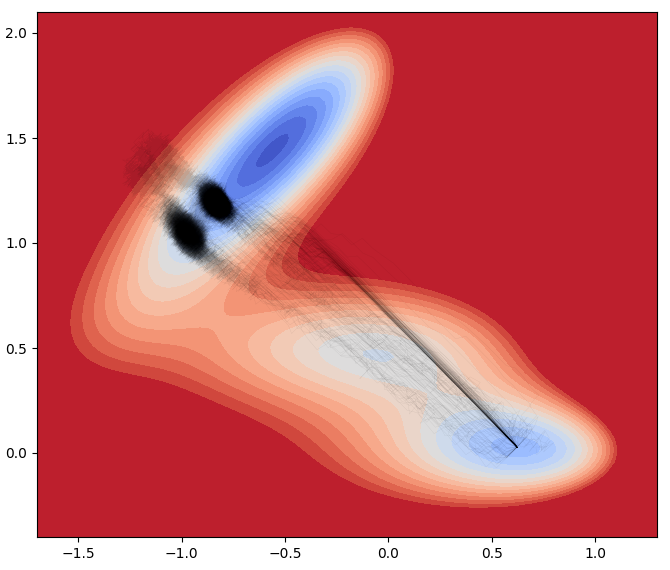}
        \caption{$\lambda = 0.02, n = 500$}
    \end{subfigure}
    \caption{Comparative scatter plots of the regions visited by the agent with different values for $\lambda$ and $n$.}
    \label{params}
\end{figure*}

A few experiments by varying the scaling factor for the action, $\lambda$, and the number of steps in an episode, $n$, were performed. The regions of the potential energy surface explored by the agent under those conditions are plotted in Figure \ref{params}. With a small value for $\lambda$, the agent does not climb out of the local minima containing the initial state (Figure \ref{params}(a)), while with a large value for $\lambda$, the agent jumps over high energy regions of the potential energy surface in the bid to reach a low energy state faster (Figure \ref{params}(f)), giving an incorrect estimate of the energy barrier for the transition.
\end{appendix}

\begin{small}
\bibliography{bibliography}
\end{small}
\end{document}